%% file: arxiv_main.tex
\documentclass[letterpaper]{article} 
\usepackage[submission]{aaai24} 
\usepackage{times} 
\usepackage{helvet}  
\usepackage{courier}  
\usepackage[hyphens]{url}  
\usepackage{graphicx} 
\usepackage{natbib} 
\usepackage{caption} 
\usepackage{algorithm}
\usepackage{algorithmic}

\usepackage{newfloat}
\usepackage{listings}

\usepackage{url}
\input{arxiv_rose}

\usepackage{caption}
\usepackage{subcaption}
\usepackage{graphicx}

\DeclareCaptionStyle{ruled}{labelfont=normalfont,labelsep=colon,strut=off} 
\floatstyle{ruled}
\newfloat{listing}{tb}{lst}{}
\floatname{listing}{Listing}

\title{Non-Parametric Representation Learning with Kernels}

\author {
    Pascal M. Esser\equalcontrib\textsuperscript{\rm 1},
    Maximilian Fleissner\equalcontrib\textsuperscript{\rm 1}
    Debarghya Ghoshdastidar\textsuperscript{\rm 1}
    }
\affiliations {
    \textsuperscript{\rm 1}Technical University of Munich,  Germany\\
    esser@cit.tum.de, fleissner@cit.tum.de, ghoshdas@cit.tum.de
}

\begin{document}

\maketitle

\begin{abstract}

Unsupervised and self-supervised representation learning has become popular in recent years for learning useful features from unlabelled data. Representation learning has been mostly developed in the neural network literature, and other models for representation learning are surprisingly unexplored. In this work, we introduce and analyze several kernel-based representation learning approaches: Firstly, we define two kernel Self-Supervised Learning (SSL) models using contrastive loss functions and secondly, a Kernel Autoencoder (AE) model based on the idea of embedding and reconstructing data. We argue that the classical representer theorems for supervised kernel machines are not always applicable for (self-supervised) representation learning, and present new representer theorems, which  show that the representations learned by our kernel models can be expressed in terms of kernel matrices. We further derive generalisation error bounds for representation learning with kernel SSL and AE, and empirically evaluate the performance of these methods in both small data regimes as well as in comparison with neural network based models.
\end{abstract}

\section{Introduction}

Representation learning builds on the idea that for most data, there exists a lower dimensional embedding that still retains most of the information usefully for a downstream task \citep{bengio2013representation}. 
While early works relied on pre-defined representations, including image descriptors such as SURF \citep{SURF} or SIFT \citep{lowe1999object} as well as bag-of-words approaches, over the past decade the focus has moved to representations learned from data itself. Across a wide range of tasks, including image classification and natural language processing \cite{bengio2013representation}, this approach has proven to be more powerful than the use of hand-crafted descriptors.
In addition, representation learning has gained increasing popularity in recent years as it provides a way of taking advantage of unlabelled data in a partially labelled data setting. Since the early works, methods for representation learning have predominantly relied on neural networks and there has been little focus on other classes of models. This may be part of the reason why it is still mostly driven from an experimental perspective.
In this work, we focus on the following two learning paradigms that both fall under the umbrella of representation learning:

\textbf{Self-Supervised representation learning  using contrastive loss functions} has been established in recent years as an important method between supervised and unsupervised learning as it does not require explicit labels but relies on implicit knowledge of what makes  samples semantically close to others. Therefore SSL builds on inputs and inter-sample relations $(\bX , \overline{\bX})$, where $\overline{\bX}$ is often constructed through data-augmentations of $\bX$ known to preserve input semantics such as additive noise or horizontal flip for an image \citep{kanazawa2016warpnet}. While the idea of SSL is not new  \citep{bromley1993signature} the main focus has been on deep SSL models, which have been highly successful in domains such as computer vision \citep{chen2020simple,Jing2019SelfSupervisedVF} and natural language processing \citep{ misra2020self,BERT2019}.

\textbf{Unsupervised representation learning through reconstruction} relies only on a set of features $\bX$ without having access to the labels. The high level idea is to map the data to a lower dimensional latent space, and then back to the features. The model is optimised by minimising the difference between the input data and the reconstruction. This has been formalized through principal component analysis (PCA) \citep{PCA} and its nonlinear extension Kernel PCA \citep{Bernhard1998neuralcomputation}.
While few approaches exist in traditional machine learning, the paradigm of representation through reconstruction has built the foundation of a large number of deep learning methods. Autoencoders (AE) \citep{kramer1991nonlinear}  use a neural network for both the embedding into the latent space as well as for the reconstruction. The empirical success of autoencoders has given rise to a large body of work, developed for task specific regularization (e.g. \citep{yang2017towards}), as well as for a wide range of applications such as image denoising \citep{buades2005review}, clustering \citep{yang2017towards} or natural language processing \cite{zhang2022survey}.
However, their theoretical understanding is still limited to analyzing critical points and dynamics in shallow linear networks \citep{Kunin20219pmlr_AE_losss_landscape, PretoriusKK18, RefinettiG22}.

\textbf{Kernel representation learning.} In spite of the widespread use of deep learning, other models are still ubiquitous in data science. For instance, decision tree ensembles are competitive with neural networks in various domains \citep{Shwartz-ZivA22,rossbach2018neural,gu2020empirical}, and are preferred due to interpretability. Another well established approach are kernel methods, which we will focus on in this paper. At an algorithmic level, kernel methods rely on the pairwise similarities between datapoints, denoted by a kernel $k(\bx,\bx')$. When the map $k$ is positive definite, $k(\bx, \bx')$ corresponds to the inner product between (potentially infinite-dimensional) nonlinear transformations of the data, and implicitly maps the data to a reproducing kernel Hilbert space (RKHS) $\calH$ through a feature map $\phi: \mathcal{X}\to\mathcal{H}$ that satisfies $k(\bx,\bx')= \langle \phi(\bx), \phi(\bx') \rangle $. Thus, any algorithm that relies exclusively on inner products can implicitly be run in $\calH$ by simply evaluating the kernel $k$. Kernel methods are among the most successful models in machine learning, particularly due to their inherently non-linear and non-parametric nature, that nonetheless allows for a sound theoretical analysis. Kernels have been used extensively in regression \citep{KIMELDORF197182,Wahba1990} and classification \citep{cortes1995support,mika1999fisher}. Since representation learning, or finding suitable features, is a key challenge is many scientific fields, we believe there is considerable scope for developing such models in these fields.
\emph{The goal of this paper is to establish that one can construct non-parametric representation learning models, based on data reconstruction and contrastive losses}. By reformulating the respective optimisation problems for such models using positive definite kernels \citep{aronszajn1950theory,ScholkopfMIT}, we implicitly make use of non-linear feature maps $\phi(\cdot)$. Moreover, the presented approaches do not reduce to traditional (unsupervised) kernel methods.
In the reconstruction based setting we define a Kernel AE and also present kernel-based self-supervised methods by considering two different contrastive loss functions.
Thereby, our work takes a significant step towards this development by decoupling the representation learning paradigm from deep learning. To this end, kernel methods are an ideal alternative since (i) kernel methods are suitable for small data problems that are prevalent in many scientific fields \citep{xu2023small,todman2023small,ChahalToner21}; (ii) kernels are non-parametric, and yet considered to be quite interpretable \citep{ponte2017kernel,hainmueller2014kernel}; and (iii) as we show, there is a natural translation from deep SSL to kernel SSL, without compromising performance.

\textbf{Contributions.} 
The main contributions of this work is the development and analysis of kernel methods for reconstruction and contrastive SSL models. More specifically:
\begin{enumerate}

\item \emph{Kernel Contrastive Learning.} 
We present \emph{kernel variants of a single hidden layer network that minimises two popular contrastive losses}. For a simple contrastive loss \citep{Arora2019ATA}, the optimisation is closely related to a kernel eigenvalue problem, while we show that the minimisation of \emph{spectral contrastive loss} \citep{HaoChen2021ProvableGF} in the kernel setting can be rephrased as a kernel matrix based optimisation.

\item \emph{Kernel Autoencoder.} We present a Kernel AE where the encoder learns a low-dimensional representation.
We show that a \emph{Kernel AE} can be learned by solving a kernel matrix based optimisation problem.

\item \emph{Theory.} We present an extension to the existing representer theorem under orthogonal constraints. 
Furthermore we derive generalisation error bounds for the proposed kernel models in  which show that the prediction of the model improve with increased number of unlabelled data.

\item \emph{Experiments.} We empirically demonstrate that the three proposed kernel methods perform on par or outperform classification on the original features as well as Kernel PCA and compare them to neural network representation learning models.

\end{enumerate}

\textbf{Related Work}
\citep{johnson2022contrastive} show that minimising certain contrastive losses can be interpreted as learning kernel functions that approximate a fixed positive-pair kernel, and hence, propose an approach of combining deep SSL with Kernel PCA. Closer to our work appears to be \citep{kiani2022joint}, where the neural network is replaced by a function learned on the RKHS of a kernel. However, their loss functions are quite different from ours. Moreover, by generalising the representer theorem, we can also enforce orthonormality on the embedding maps from the RKHS itself. \citep{zhai2023understanding} studies the role of augmentations in SSL through the lense of the RKHS induced by an augmentation. \citep{shah2022max} present a margin maximisation approach for contrastive learning that can be solved using kernel support vector machines. Their approach is close to our simple contrastive loss method (Definition~\ref{def: Contrastive simple}), but not the same as we obtain a kernel eigenvalue problem. While \citep{johnson2022contrastive,shah2022max} consider specific contrastive losses, we present a wider range of kernel SSL models, including Kernel AE, and provide generalisation error bounds for all proposed models. 

\textbf{Notation}
We denote matrices by bold capital letters $\bA$, vectors as $\ba$, and $\bI_m$ for an identity matrix of size $m \in \mathbb{N}$. For a given kernel $k: \mathbb{R}^d \times \mathbb{R}^d \rightarrow \mathbb{R}$, we denote $\phi:\bbR^d\rightarrow\calH$ for its canonical feature map into the associated RKHS $\calH$. Given data $\bx_1,\dots,\bx_n$ collected in a matrix $\bX \in \mathbb{R}^{d \times n}$, we write $\Phi:= \left(\phi(\bx_1),\dots,\phi(\bx_n)\right)$ and define $\calH_X$ as the finite-dimensional subspace spanned by $\Phi$. Recall that $\calH$ can be decomposed as $\calH_X \oplus \calH_X^\perp$. We denote by $\bK = \Phi^T\Phi\in\bbR^{n\times n}$ the kernel matrix, and define $k(\bx',\bX) = \Phi^T \phi(\bx')$. Throughout the paper we assume $n$ datapoints are used to train the representation learning model, which embeds from $\mathbb{R}^d$ into a $h$-dimensional space. On a formal level, the problem could be stated within the generalised framework of matrix-valued kernels $K: \mathbb{R}^d \times \mathbb{R}^d \rightarrow \mathbb{R}^{h \times h}$, because the vector-valued RKHS $\mathcal{H}(K)$ associated with a matrix-valued kernel $K$ naturally contains functions $\bW$ that map from $\mathbb{R}^d$ to $\mathbb{R}^h$. For the scope of this paper however, it is sufficient to assume $K(x,y) = \bI_h \cdot k(x,y)$ for some scalar kernel $k(x,y)$ with real-valued RKHS $\mathcal{H}$. Then, the norm of any $\bW \in \mathcal{H}(K)$ is simply the Hilbert-Schmidt norm that we denote as $\norm{\bW} = \norm{\bW}_\calH$ (for finite-dimensional matrices, the Frobenius norm), and learning the embedding from $\mathbb{R}^d$ to $\mathbb{R}^h$ reduces to learning $h$ individual vectors $\bw_1, \dots, \bw_h \in \mathcal{H}$. In other words, we can interpret $\bW \in \mathcal{H}(K)$ as a (potentially infinite-dimensional) matrix with columns $\bw_1, \dots, \bw_h \in \mathcal{H}$, sometimes writing $\bW = (\bw_1, \dots, \bw_h)$ for notational convenience. To underline the similarity with the deep learning framework, we denote $\bW^T \phi(x) = \left( \langle \bw_t, \phi(x) \rangle \right)_{t=1}^h = \left( \bw_1(x), \dots, \bw_h(x) \right) \in \mathbb{R}^h$, where we invoke the reproducing property of the RKHS $\mathcal{H}$ in the last step. We denote by $\bW^*$ the adjoint operator of $\bW$ (in a finite-dimensional setting, $\bW^*$ simply becomes the transpose $\bW^T$). The constraint $\bW^* \bW = \bI_h$ enforces orthonormality between all pairs $(\bw_i, \bw_j)_{i,j \le h}$.

\section{Representer Theorems}

In principle, kernel methods minimise a loss functional $\loss$ over the entire, possibly infinite-dimensional RKHS. It is the celebrated representer theorem \cite{KIMELDORF197182,ScholkopfHS01} that ensures the practical feasibility of this approach: Under mild conditions on the loss $\loss$, the optimiser is surely contained within the finite-dimensional subspace $\calH_X$. For example, in standard kernel ridge regression, the loss functional $\loss$ is simply the regularized empirical squared error
\begin{align*}
    \loss(\bw) = \sum_{i=1}^n \left(\bw(x_i) - y_i \right)^2 + \lambda \|\bw\|
\end{align*}
The fact that all minimisers of this problem indeed lie in $\calH_X$ can be seen by simply decomposing $\calH = \calH_X \oplus \calH_X^\perp$, observing that $\bw(x_i) = 0$ for all $\bw \in \calH_X^\perp$, and concluding that projecting any $\bw$ onto $\calH$ can only ever decrease the functional $\loss$. This very argument can be extended to representation learning, where regularization is important to avoid mode collapse. We formally state the following result.

\begin{theorem}(Representer Theorem for Representation Learning)
 Given data $\bx_1, \dots, \bx_n$, denote by $\loss_X(\bw_1, \dots, \bw_h)$ a loss functional on $\calH^h$ that does not change whenever $\bw_1, \dots, \bw_h$ are projected onto the finite-dimensional subspace $\calH_X$ spanned by the data. Then, any minimiser of the regularized loss functional
 \begin{align*}
    \loss(\bw_1, \dots, \bw_h) = \loss_X(\bw_1, \dots, \bw_h) + \lambda \|\bW\|_\calH
\end{align*}
consists of $\bw_1, \dots, \bw_h \in \calH_X$.
\end{theorem}

This justifies the use of kernel methods when the norm of the embedding map is penalized. However, it does not address loss functionals $\loss$ that instead impose an orthonormality constraint on the embedding $\bW$. It is natural to ask when a representer theorem exist for these settings as well. Below, we give a necessary and sufficient condition.

\begin{theorem}[\textbf{Representer theorem under orthonormality constraints}]\label{th: Representer theorem ortho}
Given data $\bX$ and an embedding dimension $h \in \mathbb{N}$, let $\loss: \mathcal{H}^h \rightarrow \mathbb{R}$ be a loss function that vanishes on $\calH^\perp_X$. Assume $\dim(\calH_X^\perp) \ge h$. Consider the following constrained minimisation problem over $\bw_1, \dots, \bw_h \in \mathcal{H}$
\begin{equation}\label{eq:orthomin1}
\begin{aligned}
    &\text{minimise } \loss(\bw_1, \dots, \bw_h) \\
    &\text{ ~ s.t. ~ } \bW^* \bW = \bI_h
\end{aligned}
\end{equation}
Furthermore, consider the inequality-constrained problem over $\calH_X$
\begin{equation}\label{eq:orthomin2}
    \begin{aligned}
    &\text{minimise } \loss(\bw_1, \dots, \bw_h) \\
    &\text{ ~ s.t. ~ } \bW^T \bW \preceq \bI_h \text{ and } \bw_1, \dots, \bw_h \in \calH_X
\end{aligned}
\end{equation}
Then, every minimiser of \eqref{eq:orthomin1} is contained in $\calH_X^h$ if and only if every minimiser of \eqref{eq:orthomin2} satisfies $\bW^T \bW = \bI_h$.
\end{theorem}

In practice, the conditions \eqref{eq:orthomin2} can often be verified directly by checking the gradient of $\loss$ on $\mathcal{H}_X$, or under orthonormalization (see Appendix). Together with the standard representer theorem, this guarantees that kernel methods can indeed be extended to representation learning --- without sacrificing the appealing properties that the representer theorem provides us with.

\section{Representation Learning with Kernels}

Building on this foundation, we can now formalize the previously discussed representation learning paradigms in the kernel setting --- namely SSL using contrastive loss functions, as well as unsupervised learning through reconstruction loss.

\subsection{Simple Contrastive Loss}\label{sec: simple contrastive Kernel Method}

For convenience, we restrict ourselves to a triplet setting with training samples $(\bx_i,\bx_i^+,\bx_i^-), {i=1,\ldots,n}$. The idea is to consider an anchor image $\bx_i$, a positive sample $\bx_i^+$ generated using data augmentation techniques, as well as an independent negative sample $\bx_i^-$. The goal is to align the anchor more with the positive sample than with the independent negative sample. In the following, we consider two loss functions that implement this idea.

In both cases, we kernelize a single hidden layer, mapping data $\bx \in \bbR^d$ to an embedding $\bz\in\bbR^h$.
\begin{align}\label{eq: contrastive Kernel mapping}
    \bx\in\bbR^{d}
    \ \xrightarrow{\phi( \cdot )}{} \ \br\in\mathcal{H}
    \  \xrightarrow{\bW}{} \ \bz\in\bbR^h.
\end{align}

We start with a simple contrastive loss inspired by \citep{Arora2019ATA}, with additional regularisation. Intuitively, this loss directly compares the difference in alignment between the anchor and the positive an the anchor and the negative sample. Formally, we define it as follows.

\begin{definition}[\textbf{Contrastive Kernel Learning}]\label{def: Contrastive simple}    
We learn a representation of the form $f_{\bW}(\bx) = \bW^T \phi(\bx)$ (see mapping in Eq.~\ref{eq: contrastive Kernel mapping}) by optimising the objective function
\begin{equation}\label{eq: simple contrastive loss}
    \begin{aligned}
        \calL^{\mathrm{Si}}&:= \sum^n_{i=1} f_{\bW}(\bx_i)^T \left(f_{\bW}(\bx_i^-) - f_{\bW}(\bx_i^+)\right)\\
    &\text{ ~ s.t. ~ } \bW^* \bW = \bI_h
    \end{aligned}
\end{equation}
\end{definition}
By verifying the conditions of Theorem~\ref{th: Representer theorem ortho}, we reduce the problem to a finite-dimensional optimisation. Theorem~\ref{th: Contrastive simple} then provides a closed from solution to the optimisation problem in Eq.~\ref{eq: simple contrastive loss}.

\begin{theorem}[\textbf{Closed Form Solution and Inference at Optimal parameterization}]\label{th: Contrastive simple}
Consider the optimisation problem as stated in Definition~\ref{def: Contrastive simple}. Let $\bX, \bX^+, \bX^- \in \bbR^{d\times n}$ denote the data corresponding to the anchors, positive and negative samples, respectively. Define the kernel matrices
\begin{align*}
    &\bK = \left[ k(x_i,x_j) \right]_{i,j}
    &&\bK_{-} = \left[ k(x_i,x_j^-) \right]_{i,j} \\
    &\bK_{+} = \left[ k(x_i,x_j^+) \right]_{i,j}
    &&\bK_{--} = \left[ k(x_i^-,x_j^-) \right]_{i,j} \\
    &\bK_{++} = \left[ k(x_i^+,x_j^+) \right]_{i,j}
    &&\bK_{-+} = \left[ k(x_i^-,x_j^+) \right]_{i,j}
\end{align*}
Furthermore, define the matrices $\bK_3 = \bK_{-} - \bK_{+}$ as well as
\begin{align*}
    &\bK_{\Delta} = \bK_{--} + \bK_{++} - \bK_{-+} - \bK_{-+}^T
    &&\bK_1 = 
    \begin{bmatrix}
    \bK & \bK_3 \\
    \bK_3 & \bK_{\Delta}  
    \end{bmatrix} \\
    &\bB = \begin{bmatrix}
    \bK_3 \\
    \bK_{\Delta} 
    \end{bmatrix} \cdot  
    \begin{bmatrix}
    \bK & \bK_{-} - \bK_{+} \\
    \end{bmatrix}
    &&\bK_2 = -\frac{1}{2} \left( \bB + \bB^T \right).
\end{align*}
Let $\bA_2$ consist of the top $h$ eigenvectors of the matrix $\bK_1^{-1/2} \bK_2 \bK_1^{-1/2}$, which we assume to have $h$ non-negative eigenvalues.
Let $\bA = \bK_1^{-1/2} \bA_2.$ Then, at optimal parameterization, the embedding of any $x^* \in \bbR^d$ can be written in closed form as
$$
\bz^*= \bA^T 
\begin{bmatrix}
    k(x^*, \bX) \\
    k(x^*, \bX^-) - k(x^*, \bX^+) 
\end{bmatrix}$$
\end{theorem}

\subsection{Spectral Contrastive Loss} \label{sec: spectral contrastive Kernel Method}
Let us now consider a kernel contrastive learning based on an alternative, commonly used spectral contrastive loss function \citep{HaoChen2021ProvableGF}.

\begin{definition}[\textbf{Spectral Kernel Learning}]\label{def: Spectral} We learn a representation of the form $f_{\bW}(\bx) = \bW^T \phi(\bx)$ (see mapping in Eq.~\ref{eq: contrastive Kernel mapping}) by optimising the following objective function, $\calL^{\mathrm{Sp}}$:
\begin{align*}
  \calL =  \sum^n_{i=1} -2 f_{\bW}(\bx_i)^T f_{\bW}(\bx_i^+) + \left( f_{\bW}(\bx_i)^T f_{\bW}(\bx_i^-) \right)^2 + \lambda \norm{\bW}^2_\calH.
\end{align*}
\end{definition}
While a closed from expression seems out of reach, we can directly rewrite the loss function using the kernel trick and optimise it using simple gradient descent. This allows us to state the following result, which yields an optimisation directly in terms of the embeddings $\bz_1, \dots, \bz_n \in \mathbb{R}^h$.

\begin{theorem}[\textbf{Gradients and Inference at Optimal Parameterization}]\label{th: Spectral}
Consider the optimisation problem as stated in Definition~\ref{def: Spectral}, with $\bK$ denoting the kernel matrix. Then, we can equivalently minimise the objective w.r.t. the embeddings $\bZ \in \bbR^{h \times 3n}$. Denoting by $\bz_1,\dots,\bz_{3n}$ the columns of $\bZ$, the loss to be minimised becomes
\begin{align*}
    &\min_{\bZ \in \bbR^{h \times 3n}} \sum^n_{i=1} -2 \bz_i^T \bz_{i+n} + \left( \bz_i^T \bz_{i+2n} \right)^2 + \lambda \cdot \Tr\left(\bZ \bK^{-1} \bZ^T \right)  
\end{align*}    
The gradient of the loss function in terms of $\bZ$ is therefore given by
\begin{align*}
 2\lambda \bZ \bK^{-1} + 
\begin{cases} 
-2 \bz_{i+n} + 2 (\bz_i^T \bz_{i+2n})  \bz_{i+2n} &, i \in [n] \\
-2 \bz_{i-n} &, i \in [n+1,2n] \\
2 (\bz_i^T \bz_{i-2n}) z_{i-2n} &, i \in [2n+1,3n]
\end{cases}
\end{align*}
For any new point $\bx^* \in \bbR^d$, the trained model maps it to
\begin{align*}
    \bz^*:=\bZ \bK^{-1}k(\bX,\bx^*).
\end{align*}
\end{theorem}

\subsection{Kernel Autoencoders}\label{sec: Kernel AE}

In general, AE architectures involve mapping the input to a lower dimensional latent space (encoding), and then back to the reconstruction (decoding). In this work we propose a Kernel AE, where both encoder and decoder correspond to kernel machines, resulting in the mapping
\begin{align*}
    {\bx}\in\bbR^{d}\xrightarrow{\phi_1( \cdot )}{} \br_1\in\mathcal{H}_1
    \xrightarrow{\bW_1}{}{\bz}\in\bbR^{h}\xrightarrow{\phi_2( \cdot )}{} \br_2\in\mathcal{H}_2
    \xrightarrow{\bW_2}{} \bx\in\bbR^{d}
\end{align*}
where typically $h<d$. While several materializations of this high-level idea come to mind, we define the Kernel AE as follows. 

\begin{definition}[\textbf{Kernel AE}]\label{def: bottleneck AE}    
Given data $ \bX\in \bbR^{d \times n}$ and a regularization parameter $\lambda>0$, define the loss functional
\begin{align*}
\calL^{AE}(\bW_1, \bW_2) := &
   \norm{\bX - \bW_2^T\phi_2\left(\bW_1^T\phi_1\left({\bX}\right)\right)}^2_\calH \\
   &+ \lambda\left(\norm{\bW_1}^2_\calH + \norm{\bW_2}^2_\calH \right)
\end{align*}
The Kernel AE corresponds to the optimisation problem
\begin{equation}\label{eq: bottleneck AE}
    \begin{aligned}
    &\min_{\bW_1, \bW_2} \calL^{AE}(\bW_1, \bW_2) \\
    &\text{ ~ s.t. ~ } \|\bW_1^T \phi( \bx_i) \|^2 = 1 ~ \forall ~ i \in [n]
    \end{aligned}
\end{equation}
\end{definition}

Let us justify our choice of architecture briefly. Firstly, we include norm regularizations on both the encoder as well as the decoder. This is motivated by the following observation: When the feature map $\phi_2$ maps to the RKHS of a universal kernel, \textbf{any} choice of $n$ distinct points $\bz_1, \dots, \bz_n$ in the bottleneck allows for perfect reconstruction. We therefore encourage the Kernel AE to learn smooth maps by penalizing the norm in the RKHS. In addition, we include the constraint $\|\bW_1^T \phi( \bx_i) \|^2 = 1 ~ \forall ~ i \in [n]$ to prevent the Kernel AE from simply pushing the points $\bz_1, \dots, \bz_n$ to zero. This happens whenever the impact of rescaling $\bz_i$ affects the norm of the encoder $\bW_1$ differently from the decoder $\bW_2$ (as is the case for commonly used kernels such as Gaussian and Laplacian). Nonetheless, we stress that other choices of regularization are also possible, and we explore some of them in the Appendix.

While a closed form solution of Definition \ref{def: bottleneck AE} is difficult to obtain, we show that the optimisation can be rewritten in terms of kernel matrices.

\begin{theorem}[\textbf{Kernel formulation and inference at optimal parameterization}]\label{th:bottleneck_kernel}
For any bottleneck $\bZ \in \mathbb{R}^{h \times n}$, define the reconstruction
\begin{align*}
    \bQ(\bZ) = \bX (\bK_Z + \lambda I_n)^{-1} \bK_Z 
\end{align*}
Learning the Kernel AE from Definition~\ref{def: bottleneck AE} is then equivalent to minimising the following expression over all possible embeddings $\bZ \in \mathbb{R}^{h \times n}$:
\begin{equation*}
    \begin{aligned}
        &\|\bQ(Z) - \bX\|^2 + \lambda \Tr\left(\bZ \bK_{ X}^{-1} \bZ^T + \bQ \bK_Z^{-1} \bQ^T \right) \\
        &\text{ ~ s.t. ~ } \|\bz_i \|^2 = 1 ~\forall i \in [n]
    \end{aligned}
\end{equation*}
Given $\bZ$, any new ${\bx}^* \in \bbR^d$ is embedded in the bottleneck as
\begin{align*}
    \bz^* = \bZ \bK_{X}^{-1} k({\bx}^*, {\bX})
    \end{align*}
and reconstructed as
\begin{align*}
    \hat{\bx}^* &= \bX \left( \bK_Z + \lambda \bI_n \right)^{-1}  k(\bz^*,\bZ)
\end{align*}
\end{theorem}

\begin{remark}[\textbf{Connection to Kernel PCA}]\label{sec: Kernel PCA}
In light of the known connections between linear autoencoders and standard PCA, it is natural to wonder how above Kernel AE relates to Kernel PCA \citep{Bernhard1998neuralcomputation}. The latter performs PCA in the RKHS $\mathcal{H}$, and is hence equivalent to minimising the reconstruction error over all orthogonal basis transformations $\bW$ in $\calH$
\begin{align}\label{eq:kernel_pca}
\loss(W) &= \sum_{i=1}^n \norm{ \phi(\bx_i) - \bW^T \bP_h \bW \phi(\bx_i) }^2
\end{align}
where $\bP_h$ denotes the projection onto the first $h$ canonical basis vectors, and we assume that the features $\phi(\bx_i)$ are centered. How does the Kernel AE $\bW_2^T \phi_2( \bW_1^T \phi_1(\bx))$ relate to this if we replace the regularisation terms on $\bW_1, \bW_2$ by an orthogonality constraint on both? For simplicity, let us assume $h=1$. The optimisation problem then essentially becomes
\begin{align}\label{eq:bottleneck_orthogonal}
\loss = \sum_{i=1}^n \norm{ \bx_i - \bW_2(\bW_1(\bx_i)) }^2
\end{align}
where $\bW_1: \bbR^d \rightarrow \bbR$ is a function from the RKHS over $\bbR^d$ (with unit norm), and $\bW_2 : \bbR \rightarrow \bbR^d$ consists of $d$ orthonormal functions from the RKHS over $\bbR$. Clearly, Eq.~\ref{eq:bottleneck_orthogonal} evaluates the reconstruction error in the sample space, much in contrast to the loss function in Eq.~\ref{eq:kernel_pca} which computes distances in the RKHS. Additionally, the map $\bW^T$ learned in Eq.~\ref{eq:kernel_pca} from the bottleneck back to $\calH$ is given by the basis transformation $\bW$ in Kernel PCA, whereas it is fixed as the feature map $\phi$ over $\bbR^h$ in the AE setting. Kernel PCA can be viewed as an AE architecture that maps solely within $\calH$, via
\begin{align*}
    \phi(\bx) \rightarrow \bW \phi(\bx) \rightarrow \bP_h \bW \phi(\bx) \rightarrow \bW^T \bP_h \bW \phi(\bx).
\end{align*}
Notably, the results of Kernel PCA usually do not translate back to the sample space easily. Given a point $\bx \in \bbR^d$, the projection of $\phi(\bx)$ onto the subspace spanned by Kernel PCA is not guaranteed to have a pre-image in $\bbR^d$, and a direct interpretation of the learned representations can therefore be difficult. In contrast, our method is quite interpretable, as it also provides an explicit formula for the reconstruction $\hat{\bx}^*$ of unseen data points --- not just their projection onto a subspace in an abstract Hilbert space. In particular, by choosing an appropriate kernel\footnote{The choice of kernel could be influenced by the type of functions that are considered interpretable in the domain of application.} and tuning the regularization parameter $\lambda$, a practitioner may directly control the complexity of both decoder as well as the encoder.

\end{remark}

\begin{remark}[\textbf{De-noising Kernel AE}]
In this section, we considered the standard setting where the model learns the reconstruction of the input data. A common extension is the \emph{de-nosing} setting (e.g. \citep{Antoni2005SIAP_denoising_survay, vincent2010stacked}), which formally moves the model from a reconstruction to a SSL setting, where we replace the input with a noisy version of the data. The goal is now to learn a function that removes the noise and, in the process, learns latent representations. More formally, the mapping becomes
\begin{align*}
\overline{\bx}\in\bbR^{d}\xrightarrow{\phi_1( \cdot )}{} \br_1\in\mathcal{H}_1
    \xrightarrow{\bW_1}{}{\bz}\in\bbR^{h}\xrightarrow{\phi_2( \cdot )}{} \br_2\in\mathcal{H}_2
    \xrightarrow{\bW_2}{} \bx\in\bbR^{d}.
\end{align*}
where $\overline{\bx}$ is given by $\overline{\bx}:=\bx + \varepsilon$ with $\varepsilon$ being the noise term. A precise formulation is provided in the Appendix. We again note that the simple extension to this setting further distinguishes our approach from Kernel PCA, where such augmentations are not as easily possible.
\end{remark}

\section{Generalisation Error Bounds}\label{sec: generalisation error bounds}
Kernel methods in the supervised setting are well established and previous works offer rigorous theoretical analysis \citep{Wahba1990,ScholkopfMIT,BartlettM02}. In this section, we show that the proposed kernel methods for contrastive SSL as well as for the reconstruction setting can be analysed in a similar fashion, and we provide generalisation error bounds for each of the proposed models. 

\subsection{Error Bound for Representation Learning Setting}

In general we are interested in characterizing
$
\calL(f)={\mathbb{E}}_{\bX \sim \mathcal{D}}\left[l\left(f(\bX)\right)\right]
$
where $f(\bX)$ is the representation function
and $l(\cdot)$ is a loss function, which is either a contrastive loss or based on reconstruction. However, since we do not have access to the distribution of the data $\calD$, we can only observe the empirical (training) error,
$
\widehat{\calL}(f)=\frac{1}{n} \sum_{i=1}^n l\left(f(\bX_i)\right)
$, where $n$ is the number of unlabelled datapoints we can characterise the generalisation error as
\begin{align*}
    \calL(f) \leq \widehat{\calL}(f) + \text{complexity term}+ \text{slack term}
\end{align*}
The exact form of the complexity and slack term depends on the embeddings and the loss. In the following, we precisely characterise them for all of the proposed models.

\begin{theorem}[\textbf{Error Bound for Kernel Contrastive Loss}]\label{th:generalisation simple contrastive}
Let $\mathcal{F}:=\left\{{\bX} \mapsto \bW^T\phi\left({\bX}\right)
:\|\bW^T\|_{\calH} \leq \omega\right\}$ be the class of embedding functions we consider in the contrastive setting. Define $\alpha:=\left(\sqrt{h\Tr\left[\bK_{\bX}\right]} + \sqrt{h\Tr\left[\bK_{\bX^-}\right]} + \sqrt{h\Tr\left[\bK_{\bX^+}\right]}\right)$ as well as $\kappa:=\max_{\bx^\prime_i\in\{\bx_i,\bx_i^-,\bx_i^+\}_{i=1}^n} k(\bx^\prime_i,\bx^\prime_i)$. We then obtain the generalisation error for the proposed losses as follows.
\begin{enumerate}
    \item 
\textbf{Simple Contrastive Loss.} Let the loss be given by Definition~\ref{def: Contrastive simple}. Then, for any $\delta>0$, the following statement holds with probability at least $1-\delta$ for any $f\in\calF$:
\begin{align*}
\calL^{\mathrm{Si}}(f)  \leq \widehat{\calL}^{\mathrm{Si}}(f) + O\left(\frac{\omega^2\sqrt{\kappa}\alpha}{n} +  \omega^2\kappa\sqrt{\frac{\log\frac{1}{\delta}}{n}}\right)
\end{align*}

\item \textbf{Spectral Contrastive Loss.} Let the loss be given by Definition~\ref{def: Spectral}. Then, for any $\delta>0$, the following statement holds with probability at least $1-\delta$ for any $f\in\calF$:
\begin{align*}
\calL^{{Sp}}(f) \leq \widehat{\calL}^{{Sp}}(f) + O\left(\lambda\omega^2 + \frac{\omega^3\kappa^\frac{3}{2}\alpha}{n} +  \omega^4\kappa^2\sqrt{\frac{\log\frac{1}{\delta}}{n}}\right)
\end{align*}
\end{enumerate} 

\end{theorem}

Similarly to the contrastive setting, we obtain a generalisation error bound for the Kernel AE as follows.

\begin{theorem}[\textbf{Error Bound for  Kernel AE}]\label{th:generalisation bottleneck}
Assume the optimisation be given by Definition~\ref{def: bottleneck AE} and define the class of encoders/decoders as: $\mathcal{F}:=\big\{{\bX} \mapsto \bW_2^T\phi_2\left(\bW_1^T\phi_1\left({\bX}\right)\right)\text{ ~ s.t. ~ } \|\bW_1^T \phi( \bx_i) \|^2 = 1 ~ \forall ~ i \in [n] ~ : ~ \|\bW_1^T\|_{\calH} \leq \omega_1,\|\bW_2^T\|_{\calH} \leq \omega_2\big\}$.
Let $r:=\lambda(\omega_1^2+\omega_2^2)$ and $\gamma = \max_{\bs\in\bbR^h\text{ ~ s.t. ~ }\norm{\bs}^2 = 1}\left\{k(\bs,\bs)\right\}$, then for any $\delta>0$, the following statement holds with probability at least $1-\delta$ for any $f\in\calF$:
\begin{align*}
    \calL^{AE}(f) \leq \widehat{\calL}^{AE}(f) + O\left(r+\frac{\omega_2\sqrt{d\gamma}}{\sqrt{n}} +   \sqrt{\frac{\log \frac{1}{\delta}}{n}}\right)
\end{align*}
\end{theorem}

The above bounds demonstrate that with increasing number of unlabelled datapoints, the complexity term in the generalisation-error bound decreases. Thus, the proposed models follow the general SSL paradigm of increasing the number of unlabelled data to improve the model performance.

\subsection{Error Bound for Supervised Downstream Task}

While the above bounds provide us with insights on the generalisation of the representation learning setting, in most cases we are also interested in the performance on downstream tasks. Conveniently, we can use the setup presented in \citep{Arora2019ATA} to bound the error of the supervised downstream tasks in terms of the unsupervised loss, providing a bound of the form
\begin{align*}
\calL_{s u p}(f) \leq c_1\widehat{\calL}_{un}(f)+c_1 * \text{complexity term}
\end{align*}
where $c_1$ and $c_2$ are data dependent constants. We present the formal version of this statement in the supplementary material for all presented models.

This highlights that a better representation (as given by a smaller loss of the unsupervised task) also improves the performance of the supervised downstream task.

\section{Experiments}\label{sec: experiments}

In this section we illustrate the empirical performance of the kernel-based representation learning models introduced in this paper. As discussed in the introduction, there is a wide range of representation learning models, that are often quite specific to the given task. We mainly consider classification in a setting with only partially labelled data at our disposal, as well as image de-noising using the Kernel AE. We state the main setup and results in the following, and provide all further details (as well as experiments on additional datasets) in the supplementary material.

\subsection{Classification on Embedding}
\textbf{Data.}
In this section, we consider the following four datasets:
\emph{concentric circles}, \emph{cubes} \citep{scikitlearn}, \emph{Iris}  \citep{fisher1936use} and \emph{Ionosphere}  \citep{misc_ionosphere_52}.
We fix the following data split: $unlabelled=50\%,~labelled = 5\%$ and $test = 45\%$, and consider $h=2$ as the embedding dimension.

\textbf{Classification task using $k$ nearest neighbours ($k$-nn) using embedding  as features.}
We investigate classification as an example of a supervised downstream task. The setting is the following: We have access to $\bX_{unlab.}$ and $\bX_{lab.}$ datapoints, which we use to train the representation learning model without access to labels. 
Then, as the downstream classification model, we consider a $k$-nn model (with $k=3$) learned on the embedding of $\bX_{lab.}$, with corresponding labels $\bY_{lab.}$. We test on $\bX_{test},\bY_{test}$.
As a benchmark, we compare to $k$-nn both on the original features as well as on the embeddings obtained by standard Kernel PCA.

\begin{figure}[t]
\centering
\includegraphics[width=0.47\textwidth]{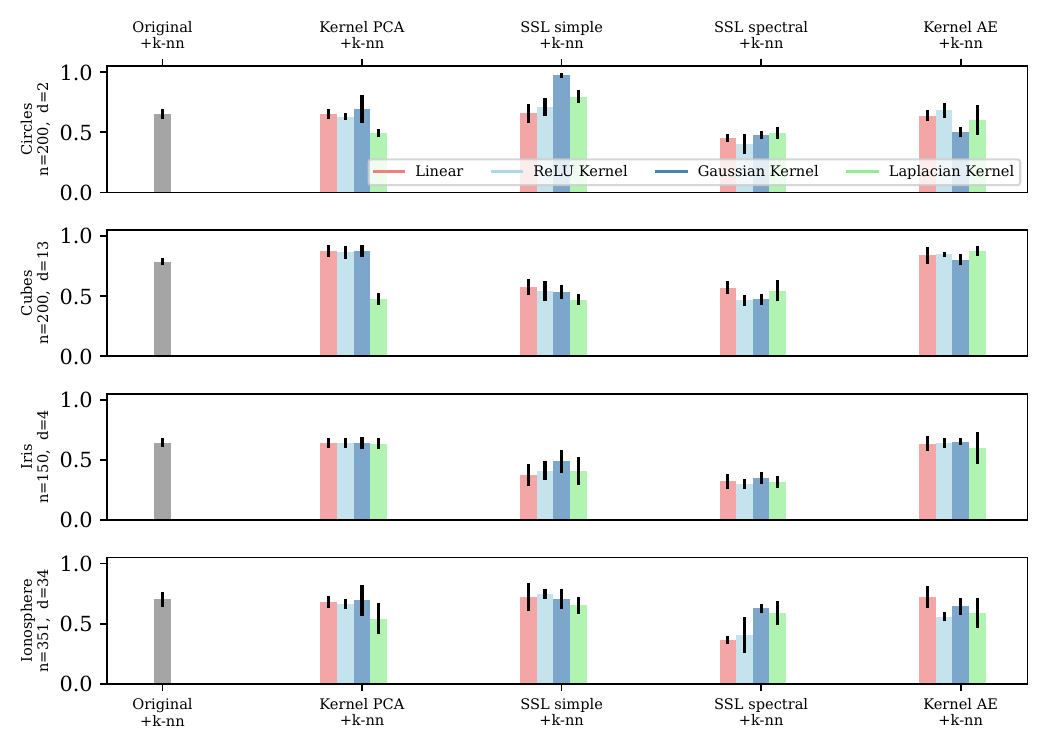}
    \caption{
    From left to right: we first consider $k$-nn on the original features followed by $k$-nn on embeddings obtained by Kernel PCA, and the proposed methods.
    }
 \label{fig: main experiments}
\end{figure}

\textbf{Choice of kernel and their parameterization.} 
For the proposed kernel methods as well as for Kernel PCA we consider three standard kernels, \emph{Gaussian, Laplacian and linear kernels} as well as a $1$-layer ReLU Kernel \citep{bietti2021ICLR}. For Gaussian and Laplacian kernel we choose the bandwidth using a grid search over $15$ steps spaced logarithmically between $0.01$ and $100$. We perform leave-one-out validation on $\bX_{lab.}$ to pick the bandwidth of the method applied to the test set.
The classification experiments on the above listed datasets are present in Figure~\ref{fig: main experiments}. All results show the mean and standard deviation over five splits of each dataset. It is apparent throughout the experiments that the choice of kernel plays a significant role in the overall performance of the model. This dependency is not surprising, as the performance of a specific kernel directly links to the underlying data-structure, and the choice of kernel is an essential part of the model design. This is in accordance with existing kernel methods --- and an important future direction is to analyze what kernel characteristics are beneficial in a representation learning setting.

\textbf{Comparison of supervised and representation learning.} As stated in the introduction (and supported theoretically in the previous section), the main motivation for representation learning is to take advantage of unlabelled data by learning embeddings that outperform the original features on downstream tasks. To evaluate this empirically for the kernel representation learning models analyzed in this paper, we compare $k$-nn on the original data to $k$-nn on the embeddings as shown in Figure~\ref{fig: main experiments}. We observe that for \emph{Circles, Cubes, Iris and Ionosphere} there always exists an embedding that outperforms $k$-nn on the original data. In addition, we observe in Figure~4 that increasing the number of unlabelled datapoints overall increases the accuracy for the downstream task as shown on the example of Linear and ReLU Kernel for the Circle dataset.

\textbf{Comparing different embedding methods.} Having observed that learning a representation before classification is beneficial, we now focus on the different embedding approaches. While the performed experiments do not reveal clear trends between different methods, we do note that the proposed methods overall perform on par or outperform Kernel PCA, underlining their relevance for kernel SSL.
 
\begin{figure}
    \centering
    \includegraphics[width = 0.47\textwidth]{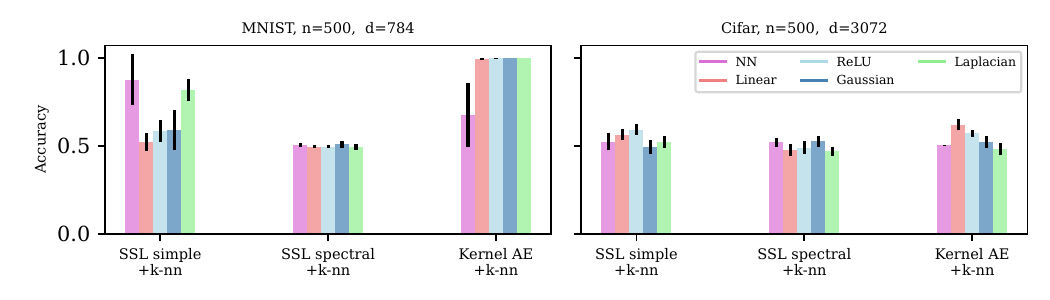}
    \caption{Comparison of kernel methods and neural network models for classification. 
    }
    \label{fig:compare NN small}
\end{figure}

\subsection{Comparison to Neural Networks for Classification and De-noising}
Representation learning has mainly been established in the context of deep neural networks. In this paper, we make a step towards decoupling the representation learning paradigm from the widely used deep learning models. Nonetheless, we can still compare the proposed kernel methods to neural networks. We construct the \emph{corresponding NN model} by replacing the linear function in the reproducing kernel Hilbert space, $\bW^\top \phi(\bx)$ by an one-hidden layer neural network $\bW_2\sigma(\bW_1\bx)$, where $\sigma( \cdot )$ is a non-linear activation function (and we still minimise a similar loss function). 

\textbf{Classification.} 
We compare the performance of both representation learning approaches in Figure~\ref{fig:compare NN small} for datasets \emph{CIFAR-10} \citep{krizhevsky2009learning}, as well as a subset of the first two classes of \emph{MNIST} \citep{deng2012mnist} (i.e. $n=500$).
We observe that the kernel methods perform on par with, or even outperform the neural networks. This indicates that there is not one dominant approach but one has to choose depending on the given task.

\begin{figure}
  \begin{subfigure}[b]{0.47\columnwidth}
    \includegraphics[width=\linewidth]{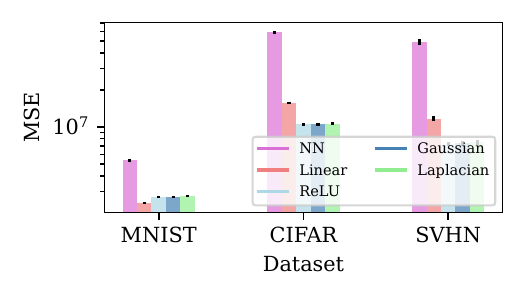}
    {Figure 3: De-noising using NN AE with and Kernel AE.}
    \label{fig:1}
  \end{subfigure}
  \hfill 
  \begin{subfigure}[b]{0.47\columnwidth}
    \includegraphics[width=\linewidth]{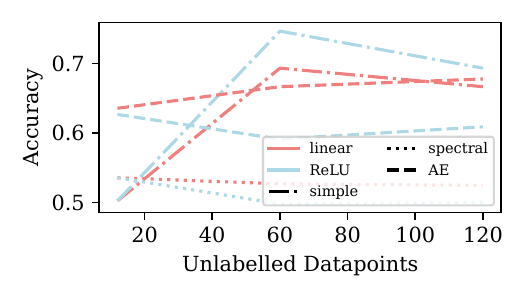}
    {Figure 4: Number of unlabelled points  (classification).}
    \label{fig:2}
  \end{subfigure}
\end{figure}

\textbf{De-Noising.} 
As a second task, we consider de-noising using (Kernel) AE. Data is sampled from the first five classes of \emph{MNIST}, \emph{CIFAR-10} and \emph{SVHN} \citep{netzer2011reading} with $n=225$ and the noisy version are generated by $\overline{\bx}:=\bx + \varepsilon, ~ \varepsilon_{i}\sim\calN(0,0.1)$. We compare the performance of kernel-based approaches with the neural network reconstructions in Figure~3 by plotting the mean square error on the test set between the AE output and the clean data. Kernel AE outperforms the neural network AE in all considered settings. Moreover, there is little variation among the different kernels. This indicates that at least in the presented settings, the proposed kernel methods pose a viable alternative to traditional neural network based representation learning.

\textbf{Formal connection between Kernel and neural network model.} While it is known that regression with infinite-width networks is equivalent to kernel regression with neural tangent kernel (NTK) \citep{Jacot2018Neurips,Arora20219NeurIPS}, similar results are not known for SSL and this brings up the question: Is kernel SSL equivalent to SSL with infinitely-wide neural networks?
 It is possible to show that single-layer Kernel AE with NTK is the infinite-width limit of over-parametrised AE \citep{Nguyen_2021_IEEE, RadhakrishnanBU20}. We believe that the same equivalence also holds for kernel contrastive learning (Definition \ref{def: Contrastive simple}) with NTK, but leave this as an open problem.
We do not know if Definition \ref{def: bottleneck AE} with NTK is the limit for bottleneck deep learning AE since, as we note earlier, there is no unique formulation for Kernel AE.

\section{Discussions and Outlook}\label{sec: related works}

In this paper, we show that new variants of representer theorem allows one to rephrase SSL optimisation problems or the learned representations in terms of kernel functions. The resulting kernel SSL models provide natural tools for theoretical analysis. 
\emph{We believe that presented theory and method provide both scope for precise analysis of SSL and can also  be extended to other SSL principles, such as other pretext tasks or joint embedding methods} \citep{Arora2019ATA,bardes2022vicreg,Grill2020Neurips,chen2020simsiam}. We conclude with some additional discussions.

\textbf{Computational limitations and small dataset setting.}
Exactly computing kernel matrices is not scalable, however random feature (RF) approximations of kernel methods are well suited for large data \citep{RahimiR07,CarratinoRR18}. While one may construct scalable kernel representation learning methods using RF, it should be noted that RF models are lazy-trained networks \citep{GhorbaniMMM19}. So fully-trained deep representation learning models may be more suitable in such scenarios.
However representation learning is relevant in all problems with availability of partially labelled data. This does not only apply to the big data regime where deep learning approaches are predominantly used, but also to \emph{small data settings} where kernel methods are traditionally an important tool \cite{fernandez2014we}.
\emph{The practical significance of developing kernel approaches is to broaden the scope of the representation learning paradigm beyond the deep learning community.}

\textbf{Kernel SSL vs. non-parametric data embedding.}
Several non-parametric generalisations of PCA, including functional PCA, kernel PCA, principle curves etc., have been studied over decades and could be compared to Kernel AEs. 
However, unlike kernel SSL, embedding methods are typically not inductive.
As shown previously, the inductive representation learning by Kernel AE and contrastive learning make them suitable for downstream supervised tasks.

\textbf{Kernel SSL vs. SSL with infinite-width neural networks.}
While it is known that regression with infinite-width networks is equivalent to kernel regression with neural tangent kernel (NTK) \cite{Jacot2018Neurips}, similar results are not known for SSL. We believe that a study of the learning dynamics of neural network based SSL would show their equivalence with our kernel contrastive models with NTK. However, it is unclear to us whether a similar result can exist for kernel AE, as NTK approximations typically do not hold in the presence of bottleneck layers \citep{0001ZB20}.

\section{Acknowledgments}
This work has been supported by the German Research Foundation (Priority Program SPP 2298, project GH 257/2-1, and Research Grant GH 257/4-1) and the DAAD programme Konrad Zuse Schools of Excellence in Artificial Intelligence, sponsored by the Federal Ministry of Education and Research.

\bibliography{arxiv_bib}

\appendix
\include{arxiv_appendix}

\end{document}

%% file: arxiv_rose.tex
\usepackage{microtype}
\usepackage{mathrsfs}
\usepackage{amsthm,amsfonts,amscd,amsmath,amssymb}
\usepackage[english]{babel}
\usepackage{mathtools}
\usepackage{libertine}
\usepackage{libertinust1math}
\usepackage{tikz}
\usepackage{xcolor}
\usepackage{microtype}
\usepackage{accents}
\usetikzlibrary{cd}
\usepackage{enumerate}
\usepackage{stmaryrd}

\usepackage{bbold}

\newcommand{\loss}{\mathcal{L}}

\DeclareMathOperator{\Tr}{Tr}

\theoremstyle{plain}
  
  \newtheorem{corollary}{Corollary}
  
  \newtheorem{lemma}{Lemma}
  \newtheorem{theorem}{Theorem}

\theoremstyle{definition}
  \newtheorem{definition}{Definition}

  \newtheorem{remark}{Remark}

\newcommand{\calD}{\mathcal{D}}

\newcommand{\calF}{\mathcal{F}}
\newcommand{\calG}{\mathcal{G}}
\newcommand{\calH}{\mathcal{H}}

\newcommand{\calL}{\mathcal{L}}

\newcommand{\calN}{\mathcal{N}}

\newcommand{\calS}{\mathcal{S}}
\newcommand{\calT}{\mathcal{T}}

\newcommand{\calX}{\mathcal{X}}

\newcommand{\bbR}{\mathbb{R}}

\newcommand{\ba}{\boldsymbol{a}}

\newcommand{\br}{\boldsymbol{r}}
\newcommand{\bs}{\boldsymbol{s}}

\newcommand{\bu}{\boldsymbol{u}}
\newcommand{\bv}{\boldsymbol{v}}
\newcommand{\bw}{\boldsymbol{w}}
\newcommand{\bx}{{\boldsymbol{x}}}
\newcommand{\by}{\boldsymbol{y}}
\newcommand{\bz}{\boldsymbol{z}}

\newcommand{\bA}{\boldsymbol{A}}
\newcommand{\bB}{\boldsymbol{B}}

\newcommand{\bI}{\boldsymbol{I}}
\newcommand{\bJ}{\boldsymbol{J}}
\newcommand{\bK}{\boldsymbol{K}}
\newcommand{\bL}{\boldsymbol{L}}

\newcommand{\bP}{\boldsymbol{P}}
\newcommand{\bQ}{\boldsymbol{Q}}

\newcommand{\bV}{\boldsymbol{V}}
\newcommand{\bW}{\boldsymbol{W}}
\newcommand{\bX}{\boldsymbol{X}}
\newcommand{\bY}{\boldsymbol{Y}}
\newcommand{\bZ}{\boldsymbol{Z}}

\newcommand{\norm}[1]{\left\lVert#1\right\rVert}

\usepackage{xargs}                      
\usepackage{todonotes}
\newcommandx{\todored}[2][1=]{\todo[linecolor=red,backgroundcolor=red!25,bordercolor=red,#1]{#2}}
\newcommandx{\todoblue}[2][1=]{\todo[linecolor=blue,backgroundcolor=blue!25,bordercolor=blue,#1]{#2}}
\newcommandx{\todogreen}[2][1=]{\todo[linecolor=green,backgroundcolor=green!25,bordercolor=green,#1]{#2}}
\newcommandx{\todoviolet}[2][1=]{\todo[linecolor=violet,backgroundcolor=violet!25,bordercolor=violet,#1]{#2}}

%% file: arxiv_appendix.tex
\onecolumn
\section*{Appendix}

~

In the supplementary material we provide the following additional proofs and results:
\begin{enumerate}[A]
    \item Model Illustration
    \item Kernel Definitions
    \item Representer Theorem
    \item Further analysis of Kernel AE
    \item Proofs for Kernel Methods at Optimal Parameterization
    \begin{enumerate}[1]        
        \item Proof Theorem~\ref{th: Spectral}
        \item Proof Theorem~\ref{th: Contrastive simple}
        \item Proof Theorem~\ref{th:bottleneck_kernel}
    \end{enumerate}
    \item{Generalisation Error Bounds}
    \begin{enumerate}[1]
    \item Proof Theorem~\ref{th:generalisation simple contrastive}
        \item Proof Theorem~\ref{th:generalisation bottleneck}
        
        \item Generalisation Error on Downstream Task
    \end{enumerate}
    \item Further Experiments 
    \begin{enumerate}[1]
        \item Further discussion and experiments comparing neural networks and kernel approach
        \item Further experiments
    \end{enumerate}
    
\end{enumerate}

~

\section{Model Illustration}
\setcounter{figure}{4}
In the following (Figure~\ref{fig:illustration}) we illustrate the main considered models: AE in the Kernel and neural network setting, Kernel PCA as well as the Kernel contrastive loss models schematically.
\begin{figure}[t]
\centering
\begin{subfigure}[b]{\textwidth}
\centering
    \includegraphics[width = \textwidth]{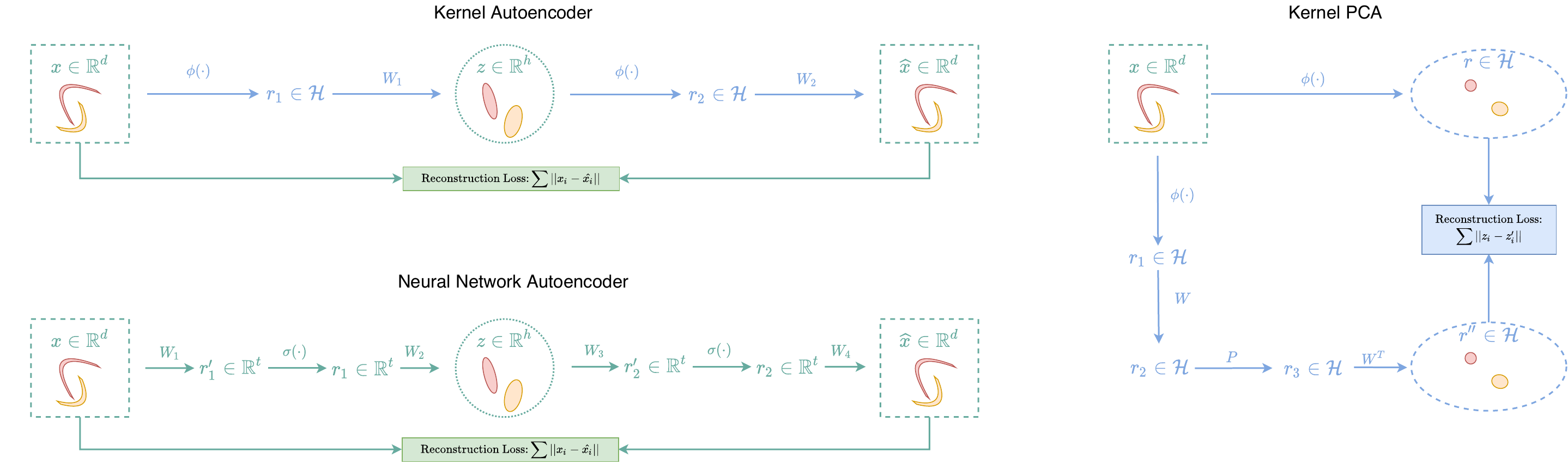}
    \caption{\textbf{Kernel AE and Kernel PCA.} Data are mapped into a lower dimensional latent space and then reconstructed. 
    \textbf{(left)} AE models: top shows the Kernel versions and bottom the traditional neural network formulation. The loss is computed between the original input $\bx$ and the reconstruction $\widehat{\bx}$.
    \textbf{(right)} Kernel PCA:
    Kernel PCA essentially performs PCA in the RKHS $\mathcal{H}$ \citep{Bernhard1998neuralcomputation} and therefore computes distances in the feature space.
We present an alternative reconstruction based formulation --- Kernel Autoencoders (KAE) --- that evaluate the reconstruction in the sample space instead of computing distances in the feature space. In comparison to Kernel PCA, this approach also provides a simple inference step for unseen datapoints.
    }
\label{fig:illustration reconstruction}
\end{subfigure}

~

\begin{subfigure}[b]{\textwidth}
\centering
    \includegraphics[width=0.55\textwidth]{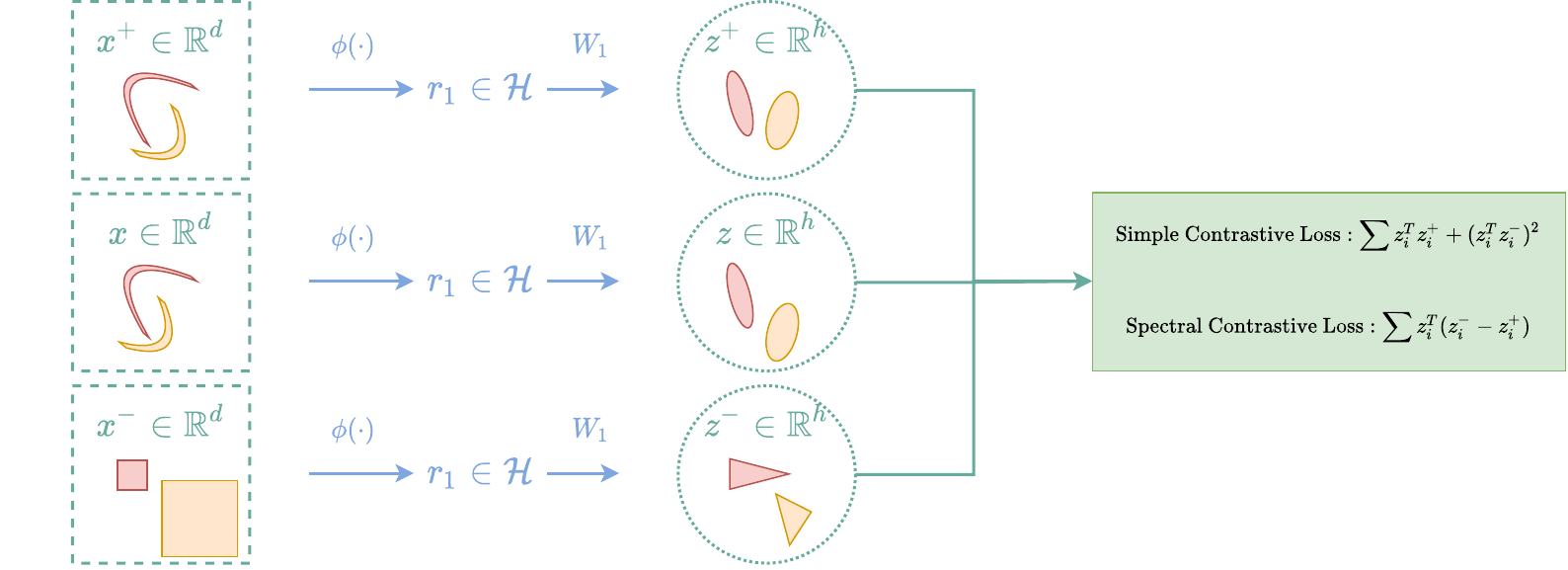}
    \caption{\textbf{Kernel Contrastive Loss.} Considering a triplet setting, the reference ($\bx$), positive ($\bx^+$) and negative ($\bx^-$) are embedded into a latent space. The function is optimised by increasing the distance between embedding $\bz,\bz^-$ and decreasing the distance between embedding of $\bz,\bz^+$. We consider two different losses, that follow this basic idea.}
\label{fig:illustration contrastive}
\end{subfigure}
\caption{Illustration of methods. Euclidean spaces and corresponding mappings are shown in green, RKHS and corresponding mappings in blue. \textbf{(a)} Overview over Reconstruction based approaches: traditional Kernel PCA, neural network based AE and Kernel AE (proposed new method).
\textbf{(b)} Proposed method for self supervised representation learning using contrastive loss functions.
}
\label{fig:illustration}
\end{figure}

\section{Kernel Definitions}\label{app: sec: Kernel Def}

For completeness we include the definitions of the kernels considered in this paper below.

\begin{definition}[Radial Basis Function (RBF) Kernel]
 Let $\bx$ and $\by$ be vectors, the the \emph{RBF kernel} is defined as:
    \begin{align*}
k(\bx, \by)=\exp \left(-\gamma\|\bx-\by\|^2\right).
\end{align*}
In the case $\gamma = \sigma^{-2}$ this becomes the Gaussian kernel of variance $\sigma^{2}$.
\end{definition}

\begin{definition}[Laplacian Kernel]
Similar to the \emph{RBF kernel}, let $\bx$ and $\by$ be vectors, then the \emph{Laplacian kernel} is defined as:
    \begin{align*}
k(\bx, \by)=\exp \left(-\gamma\|\bx-\by\|_1\right).
\end{align*}
\end{definition}

\begin{definition}[Linear Kernel]  Let $\bx$ and $\by$ be vectors, then the \emph{linear kernel} is defined as:
    \begin{align*}
k(\bx, \by)=\bx^{\top} \by.
\end{align*}
\end{definition}

In addition to the above definitions that we consider in this paper the following definition illustrate that NTK inspired kernels such as the \emph{ReLU Kernel} are also viable options to be considered as a 'pluck in option' to the proposed models.
\begin{definition}[ReLU Kernel \cite{bietti2021ICLR}]\label{def: relu kernel}
For a ReLU network with $L$ layers with inputs on the sphere, taking appropriate limits on the widths, one can show: $k_{\mathrm{NTK}}\left(\bx, \by\right)=\kappa_{\mathrm{NTK}}^L\left(\bx^{\top} \by\right)$, with $\kappa_{\mathrm{NTK}}^1(\bu)=\kappa^1(\bu)=\bu$ and for $\ell=2, \ldots, L$
\begin{align*}
\kappa^{\ell}(\bu) & =\kappa_1\left(\kappa^{\ell-1}(\bu)\right) \\
\kappa_{\mathrm{NTK}}^{\ell}(\bu) & =\kappa_{\mathrm{NTK}}^{\ell-1}(\bu) \kappa_0\left(\kappa^{\ell-1}(\bu)\right)+\kappa^{\ell}(\bu)
\end{align*}
where
\begin{align*}
\kappa_0(u)=&\frac{1}{\pi}(\pi-\arccos (u))\\
 \kappa_1(u)=&\frac{1}{\pi}\left(u \cdot(\pi-\arccos (u))+\sqrt{1-u^2}\right) .
\end{align*}
    
\end{definition}

\section{Representer theorem}\label{app: sec: reproducing kernels}
For convenience let us start by restating the theorem.

Given data $\bX$ and an embedding dimension $h \in \mathbb{N}$, let $\loss: \mathcal{H}^h \rightarrow \mathbb{R}$ be a loss function that vanishes on $\calH^\perp_X$. Assume $\dim(\calH_X^\perp) \ge h$. Consider the following constrained minimisation problem over $\bw_1, \dots, \bw_h \in \mathcal{H}$
\begin{equation}\label{eq:orthomin12}
\begin{aligned}
    &\text{minimise } \loss(\bw_1, \dots, \bw_h) \\
    &\text{ ~ s.t. ~ } \bW^* \bW = \bI_h
\end{aligned}
\end{equation}
Furthermore, consider the inequality-constrained problem over $\calH_X$
\begin{equation}\label{eq:orthomin22}
    \begin{aligned}
    &\text{minimise } \loss(\bw_1, \dots, \bw_h) \\
    &\text{ ~ s.t. ~ } \bW^T \bW \preceq \bI_h \text{ and } \bw_1, \dots, \bw_h \in \calH_X
\end{aligned}
\end{equation}
Then, every minimiser of \eqref{eq:orthomin12} is contained in $\calH_X^h$ if and only if every minimiser of \eqref{eq:orthomin22} satisfies $\bW^T \bW = \bI_h$.

Let us prove our characterization of loss functionals that admit representer theorems under orthonormality constraints. Denote $\bP_X: \calH \rightarrow \calH_X$ for the projection onto the finite-dimensional subspace $\calH_X$. Recall that $\bP_X$ is a bounded linear operator with operator norm $\| \bP_X \| =1$, satisfying $\bP_X^2 = \bP_X$.

\begin{proof}
    Let us begin by assuming that a collection of functions $\bw_1, \dots, \bw_h$ is a minimiser of
    \begin{equation}\label{eq:orthomin1_proof}
    \begin{aligned}
        &\text{minimise } \loss(\bw_1, \dots, \bw_h) \\
        &\text{ ~ s.t. ~ } \bW^* \bW = \bI_h
    \end{aligned}
    \end{equation}
    that is not contained in $\calH_X$. Then, $\bP_X \bw_1, \dots, \bP_X \bw_h \in \calH_X$ achieve the same minimum, because $\loss$ vanishes outside of $\calH_X$. Because not all $\bw_1, \dots, \bw_h$ are contained in $\calH_X$, $\bW^* \bW \neq \bI_h$. Thus, we have found a minimiser of
    \begin{equation}\label{eq:orthomin2_proof}
    \begin{aligned}
        &\text{minimise } \loss(\bw_1, \dots, \bw_h) \\
        &\text{ ~ s.t. ~ } \bW^T \bW \preceq \bI_h \text{ and } \bw_1, \dots, \bw_h \in \calH_X
    \end{aligned}
    \end{equation}
    that does not satisfy the constraint with equality. For the other direction, assume that there exists a minimiser $(\bw_1, \dots, \bw_h) \in \calH_X$ of \eqref{eq:orthomin2_proof} that does not satisfy the constraint with equality. Then, exploiting the fact that $\dim(\calH_X^\perp) \ge h$, we can simply add $\bv_1, \dots, \bv_h \in \calH_X^\perp$ to $\bw_1, \dots, \bw_h$ without changing $\loss$, while at the same time ensuring $\langle \bw_i + \bv_i , \bw_j + \bv_j \rangle = 1_{i=j}$. This new minimiser of $\loss$ is not contained in $\calH_X$ and hence there is no representer theorem.
\end{proof}

\begin{remark}
As mentioned in the main paper, checking that optimisers $\bW$ of \eqref{eq:orthomin2_proof} are indeed orthonormal can be done by analyzing
the behaviour of the loss functional $\loss$ under orthonormalization of a given solution $\bW$ with $\bW^T \bW \neq \bI_h$. To illustrate this briefly in a finite-dimensional setting, consider the trace loss
\begin{align*}
    \loss(\bW) = - \Tr(\bW^T \bA \bW)
\end{align*}
for some $\bA \succeq 0$. If $\bW^T \bW \preceq \bI_h$, then we can orthonormalize it via $\bX = \bW \bV^{-1}$, where $\bV \bV^T = \bW^T \bW$. Then, $\bX^T \bX = \bI_h$ and 
\begin{align*}
    \loss(\bX) = \Tr(\bW^T \bA \bW (\bV \bV^T)^{-1}) = \Tr(\bW^T \bA \bW (\bW^T \bW)^{-1}) \le \loss(\bW)
\end{align*}
where the final inequality follows from the fact that $\bW^T \bW \neq \bI_h$.
\end{remark}

\section{Further analysis of Kernel AE}

\textbf{Alternative formulations for the Kernel AE.}
As the overall idea of the paper is to translate deep SSL models to a kernel setting, we note at this point that there can be several ways to materialize this high level idea. Let us return to the 'bottleneck AE'. As mentioned in the main paper, when the chosen kernel is universal and $\lambda = 0$, any arbitrary choice of $n$ distinct (even one-dimensional!) points $\bz_1, \dots, \bz_n$ in the bottleneck achieves zero loss, in the sense that $\bQ = \bX$. This is in contradiction to the idea of finding a meaningful embedding of the data into lower dimensions, calling for some kind of regularization. Instead of the regularization of both $\bW_1$ as well as $\bW_2$, one could also consider the optimisation
\begin{align*}
   &\min_{\bZ, \bW} \norm{\bX - 
   \bW^T\phi(\bZ)}^2
   + \lambda  \cdot \norm{\bW}^2 \text{ ~ s.t. ~ } \|\bz_i \|^2 = 1 \forall i \in [n]
\end{align*}
which searches for a lower-dimensional embedding $\bZ$ that maps smoothly to a given $\bX$, by means of some $\bW \in \calH^{d}$. Moreover, an interesting connection to other kernel-based methods arises when we force the autoencoder to actually perfectly reconstruct the training data, i.e. when $\bQ= \bZ$. Then, the minimisation problem from Theorem \ref{th:bottleneck_kernel} reduces to the problem
\begin{equation*}
    \begin{aligned}
        &\min_{\bZ} \Tr\left(\bZ \bK_{ X}^{-1} \bZ^T + \bX \bK_Z^{-1} \bX^T \right) \\
        &\text{ ~ s.t. ~ } \|\bz_i \|^2 = 1 ~\forall i \in [n]
    \end{aligned}
\end{equation*}
Using the cyclic property of the trace, this becomes
\begin{equation*}
    \begin{aligned}
        &\min_{\bZ} \Tr\left(\bZ^T \bZ \bK_{ X}^{-1} \right) + \Tr \left( \bX^T \bX \bK_Z^{-1}\right) \\
        &\text{ ~ s.t. ~ } \|\bz_i \|^2 = 1 ~\forall i \in [n]
    \end{aligned}
\end{equation*}
which allows us to recognize $\bZ^T \bZ =: \bL_Z$ and $\bX^T \bX =: \bL_X$ as kernel matrices for $\bZ$ and $\bX$ under the linear kernel $l(s,t) = \langle s, t \rangle$. Assuming for the moment that data is centered, above expression above can be viewed as simultaneously maximizing the Hilbert-Schmidt Independence Criterion \cite{gretton2005measuring} between $\bL_X$ and $\bK_Z$, as well as $\bK_X$ and $\bL_Z$.

\section{Proofs for Kernel Methods at Optimal Parameterization}
In this section we prove the kernel expressions at optimal parameterization.

\subsection{Proof Theorem~\ref{th: Contrastive simple}}
For convenience let us start by restating the theorem.

Consider the optimisation problem as stated in Definition~\ref{def: Contrastive simple}. Let $\bX, \bX^+, \bX^- \in \bbR^{d\times n}$ denote the matrices corresponding to the anchors, positive and negative samples, respectively. Define the kernel matrices

\begin{align*}
    &\bK = \left[ k(x_i,x_j) \right]_{i,j}
    &&\bK_{-} = \left[ k(x_i,x_j^-) \right]_{i,j} \\
    &\bK_{+} = \left[ k(x_i,x_j^+) \right]_{i,j}
    &&\bK_{--} = \left[ k(x_i^-,x_j^-) \right]_{i,j} \\
    &\bK_{++} = \left[ k(x_i^+,x_j^+) \right]_{i,j}
    &&\bK_{-+} = \left[ k(x_i^-,x_j^+) \right]_{i,j}
\end{align*}

Furthermore, define the matrices

\begin{align*}
    &\bK_{\Delta} = \bK_{--} + \bK_{++} - \bK_{-+} - \bK_{-+}^T
    &&\bK_1 = 
    \begin{bmatrix}
    \bK & \bK_{-} - \bK_{+} \\
    \bK_{-} - \bK_{+} & \bK_{\Delta}  
    \end{bmatrix} \\
    &\bB = \begin{bmatrix}
    \bK_{-} - \bK_{+} \\
    \bK_{\Delta} 
    \end{bmatrix} \cdot  
    \begin{bmatrix}
    \bK & \bK_{-} - \bK_{+} \\
    \end{bmatrix}
    &&\bK_2 = -\frac{1}{2} \left( \bB + \bB^T \right)
\end{align*}

Let $\bA_2$ consist of the top $h$ eigenvectors of the matrix $\bK_1^{-1/2} \bK_2 \bK_1^{-1/2}$, and let $\bA = \bK_1^{-1/2} \bA_2.$
Then at optimal parameterization, the embedding of any $x^* \in \bbR^d$ can be written in closed form as
\begin{align*}
\bz^*= \bA^T 
\begin{bmatrix}
    k(x^*, \bX) \\
    k(x^*, \bX^-) - k(x^*, \bX^+) 
\end{bmatrix}
\end{align*}

\begin{proof}
Define the (possibly infinite-dimensional) matrices $\Phi = [\phi(x_1), \dots , \phi(x_n)]$ and
$\Delta = [\phi(x_1^-) - \phi(x_1^+), \dots , \phi(x_n^-) - \phi(x_n^+)]$, where $\phi: \bbR^d \rightarrow \calH$ is the canonical feature map associated with the given kernel and $\calH$ is the corresponding RKHS. We start by deriving the optimal parameterization. To do so recall the loss problem setup:
\begin{equation}\label{eq:simplecontrastive}
    \begin{aligned}
   &\min_{\bW} \sum^n_{i=1} f(x_i)^T \left(f(x_i^-) - f(x_i^+)\right)\\
    &\text{ ~ where ~ } f(x_i) = \bW^T \phi(x_i)\text{ ~ s.t. ~ } \bW^* \bW = \bI_h
\end{aligned}
\end{equation}
By virtue of the representer theorem under orthonormality constraints, we may reduce this to a finite-dimensional optimisation problem on the span of $(\Phi, \Delta)$, change the constraints to $\bW^T \bW \preceq I_h$ for the moment, and finally verify that the optimal solution does in fact satisfy $ \bW^T\bW = \bI_h$. If that is the case, we know that this solution is also the minimiser of \eqref{eq:simplecontrastive} over the entire space $\calH$. Hence, let us assume that there exists $\bA \in \bbR^{2n \times h}$ such that
\begin{align*}
    \bW = [\Phi, \Delta] \bA
\end{align*}
Thus, denoting $\bK_1 = [\Phi, \Delta]^T [\Phi, \Delta] \in \bbR^{2n \times 2n}$, we may rewrite our optimisation problem as
\begin{align} \label{eq:simple_contrastive_A}
   &\min_{\bA} \sum^n_{i=1} \phi(x_i)^T [\Phi, \Delta] \bA \bA^T [\Phi, \Delta]^T \left( \phi(x_i^-) - \phi(x_i^+) \right)  \\
    &\text{ ~ s.t. ~ } \bA^T \bK_1 \bA \preceq \bI_h
\end{align}
This is equivalent to
\begin{align*}
   &\min_{\bA} \Tr \left(\bA^T [\Phi, \Delta]^T \Delta \Phi^T [\Phi, \Delta] \bA \right)  \\
    &\text{ ~ s.t. ~ } \bA^T \bK_1 \bA \preceq \bI_h
\end{align*}
Denoting $\bB = [\Phi, \Delta]^T \Delta \Phi^T [\Phi, \Delta]$ and $\bK_2= -\frac{1}{2} \left( \bB + \bB^T \right)$ for the negative symmetric part of $\bB$, we are left with the trace maximization
\begin{align*}
   &\max_{\bA} \Tr \left(\bA^T \bK_2 \bA \right)  \\
    &\text{ ~ s.t. ~ } \bA^T \bK_1 \bA \preceq \bI_h
\end{align*}
Writing $\bA_2 =  \bK_1^{1/2} \bA \in \bbR^{2n \times h}$, this simplifies to
\begin{align*}
   &\max_{\bA_2} \Tr \left(\bA_2^T \bK_1^{-1/2} \bK_2 \bK_1^{-1/2} \bA_2 \right)  \\
    &\text{ ~ s.t. ~ } \bA_2^T \bA_2 \preceq \bI_h
\end{align*}
This expression is maximized when $\bA_2$ consists of the top $h$ orthonormal eigenvectors of $\bK_1^{-1/2} \bK_2 \bK_1^{-1/2}$, which do in fact satisfy  $\bA_2^T \bA_2 = \bI_h$, and hence $\bW = [\Phi, \Delta] \bA$ also satisfies $\bW^T \bW = \bI_h$ with equality. Note that both $\bK_1$ and $\bK_2$ depend only on inner products in the RKHS and can hence be computed directly from the kernel $k$, and that there is no need for evaluating the feature map directly.
Finally, $\bA = \bK_1^{-1/2} \bA_2$ becomes the desired minimiser of \ref{eq:simple_contrastive_A}. For a new point $\bx^*$, it holds that
\begin{align*}
  f(\bx^*) &= \bW^T \phi(\bx^*) \\
  &= \bA^T [\Phi, \Delta]^T \phi(\bx^*) \\
  &= \bA^T 
\begin{bmatrix}
    k(\bx^*, \bX) \\
    k(\bx^*, \bX^-) - k(\bx^*, \bX^+) 
\end{bmatrix}
\end{align*}
which again requires only knowledge of the kernel, and not of the (implicit) feature map.
\end{proof}

\subsection{Proof Theorem~\ref{th: Spectral}}
For convenience let us start by restating the theorem.

Consider the optimisation problem as stated in Definition~\ref{def: Spectral}, with $\bK$ denoting the kernel matrix. Then, we can equivalently minimise the objective w.r.t. the embeddings $\bZ \in \bbR^{h \times 3n}$. Denoting by $\bz_1,\dots,\bz_{3n}$ the columns of $\bZ$, the loss to be minimised becomes
\begin{align*}
    &\min_{\bZ \in \bbR^{h \times 3n}} \sum^n_{i=1} -2 \bz_i^T \bz_{i+n} + \left( \bz_i^T \bz_{i+2n} \right)^2 + \lambda \cdot \Tr\left(\bZ \bK^{-1} \bZ^T \right)  
\end{align*}    
The gradient of the loss function in terms of $\bZ$ is therefore given by
\begin{align*}
 2\lambda \bZ \bK^{-1} + 
\begin{cases} 
-2 \bz_{i+n} + 2 (\bz_i^T \bz_{i+2n})  \bz_{i+2n} &, i \in [n] \\
-2 \bz_{i-n} &, i \in [n+1,2n] \\
2 (\bz_i^T \bz_{i-2n}) z_{i-2n} &, i \in [2n+1,3n]
\end{cases}
\end{align*}
For any new point $\bx^* \in \bbR^d$, the trained model maps it to
\begin{align*}
    \bz^*:=\bZ \bK^{-1}k(\bX,\bx^*).
\end{align*}

\begin{proof}
Recall that in contrastive learning with the spectral contrastive loss, we learn a representation of the form $f_{\bW}(\bx) = \bW^T \phi(\bx) = \left( \bw_1(x), \dots, \bw_h(x) \right)$ by optimising the following objective function:
\begin{align*}
   \calL^{\mathrm{Sp}}:=
   \sum^n_{i=1} -2 f_{\bW}(\bx_i)^T f_{\bW}(\bx_i^+) + \left( f_{\bW}(\bx_i)^T f_{\bW}(\bx_i^-) \right)^2 + \lambda \norm{\bW}^2_\calH.
\end{align*}
Since the term $\sum^n_{i=1} -2 f_{\bW}(\bx_i)^T f_{\bW}(\bx_i^+) + \left( f_{\bW}(\bx_i)^T f_{\bW}(\bx_i^-) \right)^2$ vanishes for any choice of $\bw_1, \dots, \bw_h \in \calH_X^\perp$, and we add a norm regularization to this objective function, it is clear by the representer theorem that any minimiser of $\calL^{\mathrm{Sp}}$ must consist of $h$ functions from $\calH_X$. We may hence write $\bw_j = \Phi \ba_j $ for some $\ba_j \in \mathbb{R}^n$. Denoting $\bZ$ for the embeddings under the map $\bW^T \phi(\bx_i)$ and $\bA$ for the matrix with columns $\ba_1, \dots, \ba_h$, we see that for any $j \in [h]$, it must hold that $\bA^T \bK = \bZ$ and hence $\bA = \bK^{-1} \bZ^T$. Thus, $\bW = \Phi \bK^{-1} \bZ^T$ and $\| \bW \|^2 = \Tr(\bZ \bK^{-1} \bK \bK^{-1} \bZ^T) = \Tr(\bZ \bK^{-1} \bZ^T)$. This allows us to reformulate the minimisation problem as an optimisation over the embedded points $\bZ$, yielding the gradients
\begin{align*}
\nabla \calL^{\mathrm{Sp}} = 
 2\lambda \bZ \bK^{-1} + 
\begin{cases} 
-2 \bz_{i+n} + 2 (\bz_i^T \bz_{i+2n})  \bz_{i+2n} &, i \in [n] \\
-2 \bz_{i-n} &, i \in [n+1,2n] \\
2 (\bz_i^T \bz_{i-2n}) z_{i-2n} &, i \in [2n+1,3n]
\end{cases}
\end{align*}
Finally, by virtue of our choice of $\bW = \Phi \bK^{-1} \bZ^T$, any new point $\bx^* \in \bbR^d$ is mapped to
\begin{align*}
    \bz^*:=\bZ \bK^{-1}k(\bX,\bx^*).
\end{align*}
\end{proof}

\subsection{Proof Theorem~\ref{th:bottleneck_kernel}}
For convenience let us start by restating the theorem.

For any bottleneck $\bZ \in \mathbb{R}^{h \times n}$, define the reconstruction
\begin{align*}
    \bQ(\bZ) = \bX (\bK_Z + \lambda I_n)^{-1} \bK_Z 
\end{align*}
Learning the Kernel AE from Definition~\ref{def: bottleneck AE} is then equivalent to minimising the following expression over all possible embeddings $\bZ \in \mathbb{R}^{h \times n}$:
\begin{equation*}
    \begin{aligned}
        &\|\bQ(Z) - \bX\|^2 + \lambda \Tr\left(\bZ \bK_{ X}^{-1} \bZ^T + \bQ \bK_Z^{-1} \bQ^T \right) \\
        &\text{ ~ s.t. ~ } \|\bz_i \|^2 = 1 ~\forall i \in [n].
    \end{aligned}
\end{equation*}
Given $\bZ$, any new ${\bx}^* \in \bbR^d$ is embedded in the bottleneck as
\begin{align*}
    \bz^* = \bZ \bK_{X}^{-1} k({\bx}^*, {\bX})
    \end{align*}
and reconstructed as
\begin{align*}
    \hat{\bx}^* &= \bX \left( \bK_Z + \lambda \bI_n \right)^{-1}  k(\bz^*,\bZ)
\end{align*}

\begin{proof}
Writing $\bZ = [z_1, \dots, z_n] \in \bbR^{h \times n}$ for the points in the bottleneck and $\bQ = [q_1, \dots, q_n] \in \bbR^{d \times n}$ for the points in the output layer, and denoting $\Phi_X, \Phi_Z, \bK_X, \bK_Z$ for the respective feature maps and kernel matrices of inputs $\bX$ and bottleneck $\bZ$, the representer theorem (with norm regularization) and the same argument as in the proof of Theorem~\ref{th: Spectral} yields that the minimum-norm $\bW_1$ and $\bW_2$ satisfying $\bW_1^T \Phi_X = \bZ$ and $\bW_2^T \Phi_Z = \bQ$ are given by
\begin{align*}
    &\bW_{1} = \Phi_X \bK_{X}^{-1} \bZ^T\\
    &\bW_{2} = \Phi_Z \bK_Z^{-1} \bQ^T
\end{align*}
Their Frobenius norms (in the infinite-dimensional case, their Hilbert-Schmidt norms) are
\begin{align*}
    &\norm {\bW_{1}}^2 = \Tr(\bZ \bK_{X}^{-1} \bZ^T)\\
     &\norm {\bW_{2}}^2 = \Tr(\bQ \bK_Z^{-1} \bQ^T)
\end{align*}
Thus, the loss function is equivalent to minimising the expression
\begin{align*}
&\min_{\bZ, \bQ} \norm{\bX - \bQ}^2 + \lambda \cdot \Tr(\bZ \bK_{X}^{-1} \bZ^T + \bQ \bK_Z^{-1} \bQ^T ) \nonumber \\
&\text{ ~ s.t. ~ } \|\bz_i \|^2 = 1 \text{ ~ for all ~ }  i \in [n]
\end{align*}
Observe that for any fixed bottleneck $\bZ$, $\Tr(\bZ \bK_X^{-1} \bZ^T)$ remains constant and the above problem reduces to a sum of $d$ kernel ridge regressions, with labels $\bX$ and observations $\bZ$. Thus, the optimal parameterization $\bW_2$ simplifies to
\begin{align*}
   \bW_2 = \Phi_Z \left( \bK_Z + \lambda \bI_n \right)^{-1} \bX^T   
\end{align*}
and directly implies the final layer
\begin{align*}
    \bQ = \bX \left( \bK_Z + \lambda \bI_n \right)^{-1} \bK_Z
\end{align*}
Learning the Kernel AE from Definition~\ref{def: bottleneck AE} is hence equivalent to minimising the following expression over all possible embeddings $\bZ \in \mathbb{R}^{h \times n}$:
\begin{equation*}
    \begin{aligned}
        &\|\bQ(Z) - \bX\|^2 + \lambda \Tr\left(\bZ \bK_{ X}^{-1} \bZ^T + \bQ \bK_Z^{-1} \bQ^T \right) \\
        &\text{ ~ s.t. ~ } \|\bz_i \|^2 = 1 ~\forall i \in [n].
    \end{aligned}
\end{equation*}
Given $\bZ$, any new ${\bx}^* \in \bbR^d$ is embedded in the bottleneck as
\begin{align*}
    \bz^* = \bZ \bK_{X}^{-1} k({\bx}^*, {\bX})
    \end{align*}
and reconstructed as
\begin{align*}
    \hat{\bx}^* &= \bX \left( \bK_Z + \lambda \bI_n \right)^{-1}  k(\bz^*,\bZ)
\end{align*}
\end{proof}

\section{Generalisation Error Bounds}

Before going into the proofs of the generalisation error bounds we recall the form of generalisation error bounds we are interested in. In general would be interested in characterizing the risk in expectation over the data
\begin{align*}
\calL(f)=\underset{\bX \sim \calD}{\mathbb{E}}\left[l\left(f(\bX)\right)\right].
\end{align*}
However since we do not have access to $\calD$ we can only obtain the empirical risk quantity 
\begin{align*}
\widehat{\calL}(f)=\frac{1}{n} \sum_{i=1}^n l\left(f(\bX_i)\right).
\end{align*}
Therefore we characterize the generalisation error as 
\begin{align}\label{eq: generalisation error general}
    \calL(f) \leq \widehat{\calL}(f) + \text{complexity term}+ \text{slack term}
\end{align}
where we will consider Rademacher based characterizations of the \emph{complexity term}.

\subsection{Proof Theorem~\ref{th:generalisation simple contrastive}}

For convenience let us start by restating the theorem.

Let $\mathcal{F}:=\left\{{\bX} \mapsto \bW^T\phi\left({\bX}\right)
:\|\bW^T\|_{\calH} \leq \omega\right\}$ be the class of embedding functions we consider in the contrastive setting. Define $\alpha:=\left(\sqrt{h\Tr\left[\bK_{\bX}\right]} + \sqrt{h\Tr\left[\bK_{\bX^-}\right]} + \sqrt{h\Tr\left[\bK_{\bX^+}\right]}\right)$ as well as $\kappa:=\max_{\bx^\prime_i\in\{\bx_i,\bx_i^-,\bx_i^+\}_{i=1}^n} k(\bx^\prime_i,\bx^\prime_i)$. We then obtain the generalisation error for the proposed losses as follows.
\begin{enumerate}
    \item 
\textbf{Simple Contrastive Loss.} Let the loss be given by Definition~\ref{def: Contrastive simple}. Then, for any $\delta>0$, the following statement holds with probability at least $1-\delta$ for any $f\in\calF$:
\begin{align*}
\calL^{\mathrm{Si}}  \leq \widehat{\calL}^{\mathrm{Si}} + O\left(\frac{\omega^2\sqrt{\kappa}\alpha}{n} +  \omega^2\kappa\sqrt{\frac{\log\frac{1}{\delta}}{n}}\right)
\end{align*}

\item \textbf{Spectral Contrastive Loss.} Let the loss be given by Definition~\ref{def: Spectral}. Then, for any $\delta>0$, the following statement holds with probability at least $1-\delta$ for any $f\in\calF$:
\begin{align*}
\calL^{{Sp}} \leq \widehat{\calL}^{{Sp}} + O\left(\lambda\omega^2 + \frac{\omega^3\kappa^\frac{3}{2}\alpha}{n} +  \omega^4\kappa^2\sqrt{\frac{\log\frac{1}{\delta}}{n}}\right)
\end{align*}
\end{enumerate}

We will proof the two bounds separately.

\begin{proof}
\textbf{Part 1. Simple Contrastive Loss.}
We start from the following Lemma, that is defined in the context of the simple contrastive loss:
\begin{lemma}[\cite{Arora2019ATA}]\label{lem: simple contrastive}
With probability at least $1-\delta$ over the training set, $\forall f \in\calF$:
\begin{align*}
    \calL_{un}\leq\widehat{\calL}_{un} + O\left(R \frac{\mathfrak{R}_c(\mathcal{F}, \mathbf{x})}{n}+B\sqrt{\frac{\log \frac{1}{\delta}}{n}}\right)
\end{align*}
where $\norm{f(\cdot)}\leq R$ and $B$ the bound on the loss function wich in the case of the simple contrastive loss can be given by $B=O(R^2)$.
\end{lemma}
Furthermore the the Rademacher complexity term in the above lemma is defined over the following definition. 
\begin{definition}[Expected Rademacher Complexity for Contrastive Setting (following \citet{Arora2019ATA})]\label{def: contrastive Rademacher}
Let our dataset be consistent of triplets $\mathcal{S}=\left\{\bx_i, \bx_i^{+}, \bx_i^{-}\right\}_{i=1}^n$ and 
    $f_{\mid \mathcal{S}}=\left(f_t\left(x_j\right), f_t\left(x_j^{+}\right), f_t\left(x_j^{-}\right)\right)_{j \in[n], t \in[h]} \in \mathbb{R}^{3 h n}$ be the restriction of and $f\in\calF$ to $S$, then we define the empirical Rademacher Complexity as
     \begin{align*}
\mathfrak{R}_c(\mathcal{F}, \mathbf{x})= \underset{{\sigma\sim\{\pm 1\}^{3nh}}}{\mathbb{E}}\left[ \sup _{f \in \mathcal{F}} \sum_{i=1}^{3dn} \sigma_{i} f_{\mid \mathcal{S}}\left(x_i\right)\right].
\end{align*}
\end{definition}
Having the setup complete we can now compute the complexity term of the in this paper considered kernel function.
In the first step we pluck in our considered model and split it up by the reference, positive and negative samples:
    \begin{align*}
\mathfrak{R}_c(\mathcal{F}, \mathbf{x})&= \underset{{\sigma\sim\{\pm 1\}^{3nd}}}{\mathbb{E}}\left[ \sup _{f \in \mathcal{F}} \sum_{i=1}^{3hn} \sigma_{i} f_{\mid \mathcal{S}}\left(x_i\right)\right]\\
&= \underset{{\sigma\sim\{\pm 1\}^{3nd}}}{\mathbb{E}}\left[ \sup _{\bW:\norm{\bW}_\calH\leq\omega} \sum_{j=1}^h\sum_{i=1}^{n} \sigma_{i} \bW_{\cdot,j}^T\phi\left({\bx_i}\right) + \sigma_{i+n} \bW_{\cdot,j}^T\phi\left({\bx^-_i}\right)+ \sigma_{i+2n} \bW_{\cdot,j}^T\phi\left({\bx^+_i}\right) \right]\\
&\leq\omega\left(\underset{\sigma}{\mathbb{E}}\left[h\norm{\sum_{i=1}^{n}\sigma_{i}\phi\left({\bx_i}\right)}\right]
+\underset{\sigma}{\mathbb{E}}\left[h\norm{\sum_{i=1}^{n}\sigma_{i}\phi\left({\bx^-_i}\right)}\right]
+\underset{\sigma}{\mathbb{E}}\left[h\norm{\sum_{i=1}^{n}\sigma_{i}\phi\left({\bx^+_i}\right)}\right]\right)\tag{by linearity of expectation and Cauchy-Schwartz inequality}\\
&=\omega\left(\underset{\sigma}{\mathbb{E}}\left[\sqrt{h\norm{\sum_{i=1}^{n}\sigma_{i}\phi\left({\bx_i}\right)}^2}\right]
+\underset{\sigma}{\mathbb{E}}\left[\sqrt{h\norm{\sum_{i=1}^{n}\sigma_{i}\phi\left({\bx^-_i}\right)}^2}\right]
+\underset{\sigma}{\mathbb{E}}\left[\sqrt{h\norm{\sum_{i=1}^{n}\sigma_{i}\phi\left({\bx^+_i}\right)}^2}\right]\right)\\
&\leq\omega\left(\sqrt{h\underset{\sigma}{\mathbb{E}}\left[\norm{\sum_{i=1}^{n}\sigma_{i}\phi\left({\bx_i}\right)}^2\right]}
+\sqrt{h\underset{\sigma}{\mathbb{E}}\left[\norm{\sum_{i=1}^{n}\sigma_{i}\phi\left({\bx^-_i}\right)}^2\right]}
+\sqrt{h\underset{\sigma}{\mathbb{E}}\left[\norm{\sum_{i=1}^{n}\sigma_{i}\phi\left({\bx^+_i}\right)}^2\right]}\right)\tag{Jensen’s inequality}\\
& = \omega\left(\sqrt{h\Tr\left[\bK_{\bX}\right]} + \sqrt{h\Tr\left[\bK_{\bX^-}\right]} + \sqrt{h\Tr\left[\bK_{\bX^+}\right]}\right).
\end{align*}
Secondly we have to bound the quantity  $\norm{f(\cdot)}\leq R$:
\begin{align*}
    \norm{f(\cdot)} & = \norm{\bW^T\phi(\bx)}\tag{by definition of considered embedding function}\\
    & \leq \norm{\bW^T}\norm{\phi(\bx)}\\
    & \leq \omega\norm{\phi(\bx)}\tag{by definition of function class $\norm{\bW^T}$ is bound}\\
    & \leq \omega\sqrt{\langle\phi(\bx)^T\phi(\bx)\rangle}\\
    &\leq\omega\sqrt{\max_{\bx_i}. k(\bx_i,\bx_i)}\tag{bounding over all possible $\bx_i$}
\end{align*}
Defining $\kappa:=\max_{\bx^\prime_i\in\{\bx_i,\bx_i^-,\bx_i^+\}_{i=1}^n} k(\bx^\prime_i,\bx^\prime_i)$ to account for reference, positive and negative samples and combining all results concludes this part of the proof.
\end{proof}

\begin{proof}
\textbf{Part 2. Spectral Contrastive Loss.}

The overall proof structure follows the one presented above for the simple contrastive loss, however Lemma~\ref{lem: simple contrastive} is define for the simple contrastive loss. Therefore we will adept the proof of \citet{Arora2019ATA} Lemma A.2. to obtain the following lemma for the spectral contrastive loss.
\begin{lemma}\label{lem: spectral}
    With probability at least $1-\delta$ over the training set, $\forall f \in\calF$:
\begin{align}\label{eq: lemma 2}
    \calL_{un}\leq\widehat{\calL}_{un} + O\left(R^3 \frac{\mathfrak{R}_c(\mathcal{F}, \mathbf{x})}{n}+B \sqrt{\frac{\log \frac{1}{\delta}}{n}}\right)
\end{align}
where $\mathfrak{R}_c(\mathcal{F}, \mathbf{x})$ is the Vector Rademacher Complexity where $\norm{f(\cdot)}\leq R$ and $B=O(R^4)$.
\end{lemma}

Before proofing Lemma~\ref{lem: spectral} we first recall the following Lemma:
\begin{lemma}[Corollary 4 in \cite{MauerRad}] Let $Z$ be any set and $\calS=\{z_j\}^n_{j=1}\in Z^n$. Let $\widetilde{\calF}$ be a class of functions  $\tilde{f}:Z\rightarrow\bbR^d$ and $h:\bbR^d\rightarrow\bbR$ be $L$-Lipschitz. For all $\tilde{f}\in\widetilde{\calF}$, let $g_{\tilde{f}} = h\circ\tilde{f}$. Then
\begin{align}\label{eq: lemma 3}
\underset{\sigma \sim\{ \pm 1\}^n}{\mathbb{E}}\left[\sup _{\tilde{f} \in \tilde{\mathcal{F}}}\left\langle\sigma,\left(g_{\tilde{f}}\right)_{\mid \mathcal{S}}\right\rangle\right] \leq \sqrt{2} L \underset{\sigma \sim\{ \pm 1\}^{d n}}{\mathbb{E}}\left[\sup _{\tilde{f} \in \tilde{\mathcal{F}}}\left\langle\sigma, \tilde{f}_{\mid \mathcal{S}}\right\rangle\right]
\end{align}
where $\tilde{f}_{\mid \mathcal{S}}=\left(\tilde{f}_t\left(z_j\right)\right)_{t \in[d], j \in[n]}$.
\end{lemma}

We start by considering the classical Rademacher complexity based generalization error. For a real function class $\calG$ whose functions map from a set $Z$ to $[0, 1]$ and for any $\delta > 0$, if $\calS$ is a training set composed by $n$ i.i.d. samples $\{z_j\}_{j=1}^n$, then with probability at least $1-\frac{2}{\delta}$,for all $g \in \calG$
\begin{align*}
\mathbb{E}[g(z)] \leq \frac{1}{n} \sum_{j=1}^n g\left(z_i\right)+\frac{2 \mathfrak{R}_{\mathcal{S}}(\calG)}{n}+3 \sqrt{\frac{\log \frac{4}{\delta}}{2 n}}
\end{align*}
where $\mathfrak{R}_{\mathcal{S}}(G)$ is the standard Rademacher complexity.
We can apply this to our setting by considering $Z=\calX^3$ and defining the function class as
\begin{align*}
\calG=\left\{g_f\left(x, x^{+}, x^{-}\right)
=\frac{1}{B}\left( f(x)^Tf\left(x^{+}\right)-\left(f\left(x_i\right)^Tf\left(x_i^{-} \right)\right)^2\right)
\mid f \in \mathcal{F}\right\}.
\end{align*}
Now to show \eqref{eq: lemma 2} consider some universal constant $c$ we have to show $\mathfrak{R}_{\mathcal{S}}(\calG)\leq c\frac{R^3}{B}\mathfrak{R}_{\mathcal{S}}(\calG)$ or equivalently
\begin{align}\label{eq: rademcher comaparison}
\underset{\sigma \sim\{ \pm 1\}^n}{\mathbb{E}}\left[\sup _{f \in \mathcal{F}}\left\langle\sigma,\left(g_f\right)_{\mid \mathcal{S}}\right\rangle\right] \leq c \frac{ R^3 }{B} \underset{\sigma \sim\{ \pm 1\}^{3d n}}{\mathbb{E}}\left[\sup _{f \in \mathcal{F}}\left\langle\sigma, f_{\mid \mathcal{S}}\right\rangle\right]
\end{align}
where $\left(g_f\right)_{\mid \mathcal{S}}=\left\{g_f\left(x_j, x_j^{+}, x_{1}^{-}\right)\right\}_{j=1}^n$.
We can now observe by setting $Z= \calX^3, b = 3d$ and 
\begin{align*}
\widetilde{\mathcal{F}}=\left\{\tilde{f}\left(x, x^{+}, x^{-}\right)=\left(f(x), f\left(x^{+}\right), f\left(x^{-}\right), \right) \mid f \in \mathcal{F}\right\}
\end{align*} and using $g_{\tilde{f}}=g_f$ 
that \eqref{eq: lemma 3} and \eqref{eq: rademcher comaparison} exactly coincide and we need to show $L\leq \frac{c}{\sqrt{2}}\frac{R^3}{B}$ for some constant $c$. Now for $z = (x,x^+,x^-)$ we have $g_{\tilde{f}}(z)=\frac{1}{B} \psi(\tilde{f}(z))$ where $\psi: \mathbb{R}^{(1+2) d} \rightarrow \mathbb{R}$ with
$
\psi\left(\left(v_t, v_t^{+}, v_{t}^{-}\right)_{t \in[d]}\right)=\sum_t -v_tv_t^{+} + \left(v_tv_{t }^{-}\right)^2.
$
We can now show that $\psi$ is $R^3$ lipschitz where $\sum_t v_t^2, \sum_t\left(v_t^{+}\right)^2, \sum_t\left(v_{t}^{-}\right)^2 \leq R^2$
by computing its Jacobian. To do so we derive
$
\frac{\partial \psi_i}{\partial v_t^{+}}=-v_t\quad 
\frac{\partial \psi_i}{\partial v_t}=-v_t^{+}+v_{t }^{-}v_{t }^{-}v_{t }
$ and $ 
\frac{\partial \psi_i}{\partial v_{t }^{-}}=v_{t }v_{t }v_{t }^{-}
$
and get by triangle inequality the Frobenius norm on the Jacobian $\bJ$ of $\psi$
$
    \norm{\bJ}_F \leq \sqrt{ \sum_t (v_{t }^{-})^4(v_{t })^2
    + (v_{t }^{-})^3 v_{t }^3
    + (v_{t }^{-})^2 v_{t }^4
    + (v_{t }^{+})^2
    + (v_{t }^{+}) v_t
    + (v_{t })^2} =O( R^3).
$
Finally using $\norm{\bJ}_2\leq\norm{\bJ}_F$ bounds the lipschitzness and concludes the proof of Lemma~\ref{lem: spectral}.

As the function class we consider for embedding does not change $(\mathcal{F}:=\left\{{\bX} \mapsto \bW^T\phi\left({\bX}\right)
:\|\bW^T\|_{\calH} \leq \omega\right\})$, we again obtain
$
    \mathfrak{R}_c(\mathcal{F}, \mathbf{x})\leq \omega\left(\sqrt{h\Tr\left[\bK_{\bX}\right]} + \sqrt{h\Tr\left[\bK_{\bX^-}\right]} + \sqrt{h\Tr\left[\bK_{\bX^+}\right]}\right)
$
and
$
    \omega\sqrt{\kappa}\leq R,
$
which combined with the above Lemma~\ref{lem: spectral} concludes the proof.
\end{proof}

\subsection{Proof Theorem~\ref{th:generalisation bottleneck}}
For convenience let us start by restating the theorem.

Assume the optimisation be given by Definition~\ref{def: bottleneck AE}and define the class of encoders/decoders as: $\mathcal{F}:=\big\{{\bX} \mapsto \bW_2^T\phi_2\left(\bW_1^T\phi_1\left({\bX}\right)\right)\text{ ~ s.t. ~ } \|\bW_1^T \phi( \bx_i) \|^2 = 1 ~ \forall ~ i \in [n] ~ : ~ \|\bW_1^T\|_{\calH} \leq \omega_1,\|\bW_2^T\|_{\calH} \leq \omega_2\big\}$.
Let $\gamma = \max_{\bs\in\bbR^h\text{ ~ s.t. ~ }\norm{\bs}^2 = 1}\left\{k(\bs,\bs)\right\}$ and $r:=\lambda(\omega_1^2+\omega_2^2)$, then for any $\delta>0$, the following statement holds with probability at least $1-\delta$ for any $f\in\calF$:
\begin{align*}
    \calL^{AE} \leq \widehat{\calL}^{AE} + O\left(r+\frac{\omega_2\sqrt{d\gamma}}{\sqrt{n}} +   \sqrt{\frac{\log \frac{1}{\delta}}{n}}\right)
\end{align*}

\begin{proof}
Following to the general form stated in Eq.~\ref{eq: generalisation error general} we have to characterize the \emph{complexity and the slack term.}
We will start by  following a version of the standard Rademacher complexity to account for the multi-dimensional output to bound the former.

\begin{definition}[Empirical Vector Rademacher Complexity following \cite{maurer2016bounds}]\label{def: vector rademacher}
Let us consider a function class $\calF:=\left\{f:\bbR^d\rightarrow\bbR^h\right\}$ and a dataset $S = \{\bx_i\}_{i=1}^n, \bx_i\in\bbR^d$. Let $I:[h]\rightarrow2^{[n]}$ be a function which assignes to every $t\in[h]$ a subset $I_t\subset[n]$ and $\sigma_{ij}$ are doubly indexed, independent Rademacher variables. Then we define the Rademacher complexity as
    \begin{align*}
\mathfrak{R}_I(\mathcal{F}, \mathbf{x})=\frac{1}{n} \underset{}{\mathbb{E}}\left[ \sup _{f \in \mathcal{F}} \sum_{t=1}^h \sum_{i \in I_t} \sigma_{t i} f_t\left(x_i\right)\right].
\end{align*}
\end{definition}
 In simple terms this is a standard Rademacher approach while taking the dimension over the output dimension into account.

We first start with the overall loss function and 
\begin{align*}
    \calL\left(\bX, f_{\bW_1, \bW_2}^{BAE}\left(\overline{\bX}\right) \right)&:= 
   \norm{\bX - \bW_2^T\phi_2\left(\bW_1^T\phi_1\left(\overline{\bX}\right)\right)}^2_\calH 
   + \lambda\left(\norm{\bW_1}^2_\calH + \norm{\bW_2}^2_\calH \right)\\ &\text{ ~ s.t. ~ } \|\bW_1^T \phi(\overline \bx_i) \|^2 = 1 ~ \forall ~ i \in [n],
\end{align*}
and use the additive nature of Rademacher complexity to bound the regularization terms first by $\omega_1,\omega_2$.
Secondly noting that the square norm is L-Lipschitz and using the Lipschitz composition property of Rademacher complexity to bound
\begin{align*}
    \mathfrak{R}_I(\ell\circ\mathcal{F}, \mathbf{x})\leq L\mathfrak{R}_I(\mathcal{F}, \mathbf{x})
\end{align*}
and we therefore can focus on the encoding-decoding function $\mathfrak{R}_I(\mathcal{F}, \mathbf{x})$.
Starting from this general formulation we can now apply this to our setting
\begin{align*}
    \mathfrak{R}_I(\mathcal{F}, \mathbf{x})
    &=\frac{1}{n} \underset{}{\mathbb{E}}\left[ \sup _{f \in \mathcal{F}} \sum_{t=1}^h \sum_{i \in I_t} \sigma_{t i} f_t\left(x_i\right)\right]\tag{by Definition~\ref{def: vector rademacher}}\\
    &=\frac{1}{n}\mathbb{E}\left[\sup_{\bW_1,\bW_2:\norm{\bW_2}_\calH \leq\omega_2}\sum_t^d\sum_{i\in I_{t}} \sigma_{ti}\bW_{\cdot t}\phi\left(\bW_1\sigma(\bx_i)\right)\right]\tag{model definition}\\
    &=\frac{1}{n}\mathbb{E}\left[\sup_{\bW_1,\bW_2:\norm{\bW_2}_\calH\leq\omega_2}\sum_t^d\left\langle \bW_{\cdot t},\sum_{i\in I_{t}} \sigma_{ti}\phi\left(\bW_1\sigma(\bx_i)\right)\right\rangle\right]\\
    &=\frac{\omega_2}{n}\mathbb{E}\left[\sup_{\bW_1}\sqrt{\sum_t^d\norm{\sum_{i\in I_{t}} \sigma_{ti}\phi\left(\bW_1\sigma(\bx_i)\right)}^2}\right]\\
    &\leq\frac{\omega_2}{n}\sqrt{\sum_t^d\mathbb{E}\left[\sup_{\bW_1}\norm{\sum_{i\in I_{t}} \sigma_{ti}\phi\left(\bW_1\sigma(\bx_i)\right)}^2\right]}\tag{Jenson inequality}\\
    &=\frac{\omega_2}{n}\sqrt{d\sum_i\sup_{\bW_1}\norm{\phi\left(\bW_1\sigma(\bx_i)\right)}^2}\tag{$\mathbb{E}\left[\sigma_i\sigma_j\right] = 0, i\neq j$}
\end{align*}
Now recall that by definition $\|\bW_1^T \phi( \bx_i) \|^2 = 1 ~ \forall ~ i \in [n]$. Therefore picking the supremum over $\bW_1$ is obtained for $\gamma = \max_{\bs\in\bbR^h\text{ ~ s.t. ~ }\norm{\bs}^2 = 1}\left\{k(\bs,\bs)\right\}$ and 
\begin{align*}
    \frac{\omega_2}{n}\sup_{\bW_1}\sqrt{d\sum_i\norm{\phi\left(\bW_1\sigma(\bx_i)\right)}^2}
    &\leq \frac{\omega_2}{n}\sqrt{dn\gamma}\\
    & = \frac{\omega_2\sqrt{d\gamma}}{\sqrt{n}}
\end{align*}

Combining the above with the standard generalisation error bound \citep{bartlett2002rademacher} in the regression setting adds the \emph{slack term} and concludes the proof.
\end{proof}

\subsection{Generalisation Error on Downstream Task}
We can use the setup presented in \citep{Arora2019ATA} to bound the supervised error of the downstream tasks by the unsupervised as computed above. Before we state the bound let us formally define the supervised task. 
We consider a two-class classification task $\calT$ with $\{c_1,c_2\}$ distinct classes and a linear classifier on top of the learned representation. Let this function given by $\bV\in\bbR^{2\times h}$. In the following let $\bx_c$ be a datapoint $\bx$ belonging to class $c$.
\begin{align*}
\calL_{\text {sup }}(\mathcal{T}, f)=\inf_{\bV}  \underset{(\bx, c)}{\mathbb{E}}\left[\bV f(\bx_{c_1})-\bV f(\bx_{c_2}) \mid c_1 \neq c_2\right]
\end{align*}
From there we can furthermore define the average supervised loss as taking the expectation over the distribution of classes.
The average loss for a function $f$ on a binary classification task tasks is defined as
    \begin{align*}
\calL_{\text {sup }}(f):=\underset{\{c_1,c_2\} \sim \rho^{2}}{\mathbb{E}}\left[L_{s u p}\left(\{c_1,c_2\}, f\right) \mid c_1 \neq c_2\right]
\end{align*}
where the latent class distribution is given by $\rho$.
From there we can now bound the supervised loss by the corresponding unsupervised one.
\begin{corollary}[\textbf{Error Bound on Downstream Tasks}] \label{cor: supervised bound}
Let $t=\underset{c, c^{\prime} \sim \rho^2}{\mathbb{E}} \mathbf{1}\left\{c=c^{\prime}\right\}$ and $\tau:=\frac{1}{(1-t)}$ be the probability that two classes sampled independently from $\rho$ are the same. Again define $\alpha:=\left(\sqrt{h\Tr\left[\bK_{\bX}\right]} + \sqrt{h\Tr\left[\bK_{\bX^-}\right]} + \sqrt{h\Tr\left[\bK_{\bX^+}\right]}\right)$. In the following let $\calL_{s u p}$ be the loss of the \emph{supervised downstream task}.
\begin{enumerate}

\item 
\textbf{Simple Contrastive Loss.}
Let $\widehat{\calL}^{Si}$ be the simple contrastive loss as defined in Definition~\ref{def: Contrastive simple}. Then for any $\delta>0$, the following statement holds with probability at least $1-\delta$ for any $f\in\calF$:
\begin{align*}
\calL_{s u p}^{\mathrm{Si}} \leq& \tau\left(\widehat{\calL}_{un}^{\mathrm{Si}}-t\right)+\tau O\left(\frac{\omega^2\sqrt{\kappa}\alpha}{n} +  \omega^2\kappa\sqrt{\frac{\log\frac{1}{\delta}}{n}}\right)
\end{align*}

\item 
\textbf{Spectral Contrastive Loss.}
Let $\widehat{\calL}^{\mathrm{Sp}}$ be the  spectral contrastive loss as defined in Definition~\ref{def: Spectral}.
For any $\delta>0$, the following statement holds with probability at least $1-\delta$ for any $f\in\calF$:
\begin{align*}
    \calL_{s u p}^{\mathrm{Sp}} \leq& \tau\left(\widehat{\calL}^{\mathrm{Sp}}_{un} -t\right)
    +\tau O\left(\frac{\omega^3\kappa^\frac{3}{2}\alpha}{n} +  \omega^6\kappa^3\sqrt{\frac{\log\frac{1}{\delta}}{n}}\right)
\end{align*}

\item \textbf{Kernel AE.}
Consider the embedding function from the function class $\mathcal{F}:=\left\{{\bX} \mapsto \bW^T\phi\left({\bX}\right)
:\|\bW^T\|_{\calH} \leq \omega\right\}$ and let be $\widehat{\calL}_{un}^{AE+}$ the loss on the embedding for $\bx^+$ and 
$\widehat{\calL}_{un}^{AE-}$ the loss on the embedding for $\bx^-$, standing in for two classes\footnote{Remark: while it seems surprising that positive and negative samples suddenly appear in the AE setup we note that in the contrastive setting this allows to naturally account for mappings to different classes. Therefore in the AE, introducing this setting allows for class differentiation in the embedding.}. Furthermore let $\Tr[\bK_{\bX^+} ], \Tr[\bK_{\bX^-} ] \leq \beta$
For any $\delta>0$, the following statement holds with probability at least $1-\delta$ for any $f\in\calF$:
\begin{align*}
\calL_{s u p}^{AE} \leq \tau\left(\left|\widehat{\calL}_{un}^{AE+} - \widehat{\calL}_{un}^{AE-}\right| -t\right) + \tau O\left(\frac{\omega\sqrt{h\beta}}{\sqrt{n}} +  \sqrt{\frac{\log \frac{1}{\delta}}{n}}\right)
\end{align*}

\end{enumerate}
\end{corollary}

This importantly highlights that according to the above bounds a better representation (as given by a smaller loss of the unsupervised task) also improves the performance of the supervised downstream task.
\begin{proof}
Before we state the bound let us formally define the supervised task. 
We consider a two-class classification task $\calT$ with $\{c_1,c_2\}$ distinct classes and a linear classifier on top of the learned representation. Let this function given by $\bV\in\bbR^{2\times h}$. In the following let $\bx_c$ be a datapoint $\bx$ belonging to class $c$.
\begin{align*}
\calL_{\text {sup }}(\mathcal{T}, f)=\inf_{\bV}  \underset{(\bx, c)}{\mathbb{E}}\left[\bV f(\bx_c)-\bV f(\bx_{c^{\prime}}) \mid c_i \neq c_j\right]
\end{align*}
From there we can furthermore define the average supervised loss as taking the expectation over the distribution of classes.
The average loss for a function $f$ on a binary classification task tasks is defined as
    \begin{align*}
\calL_{\text {sup }}(f):=\underset{\{c_1,c_2\} \sim \rho^{2}}{\mathbb{E}}\left[L_{s u p}\left(\{c_1,c_2\}, f\right) \mid c_i \neq c_j\right]
\end{align*}
where the latent class distribution is given by $\rho$.

In the following let $\mu_c=\underset{x \sim \calD_c}{\mathbb{E}} f(x)$ be the mean of class $c$ and $f(\bx)$ the embedding function.
\begin{enumerate}
    \item 

\textbf{Simple Contrastive Loss.} This lemma is directly proven for the simple contrastive loss in \citet{Arora2019ATA}. We will first restate the proof for the simple contrastive loss from completeness.  We can now bound the unsupervised loss:

\begin{align*}
\calL_{u n}(f)&=\underset{\substack{\left(\bx, \bx^{+}\right) \sim \calD_{\text {sim }} \\ \bx^{-} \sim \calD_{\text {neg }}}}{\mathbb{E}}\left[f(\bx)^T\left(f\left(\bx^{+}\right)-f\left(\bx^{-}\right)\right)\right]    \\
&=
\underset{\substack{c^{+}, c^{-} \sim \rho^2 \\
x \sim \calD_{c^{+}}}}{\mathbb{E}}
\underset{\substack{
x^{+} \sim \calD_{c^{+}} \\
x^{-} \sim \calD_{c^{-}}}}{\mathbb{E}}\left[f(\bx)^T\left(f\left(\bx^{+}\right)-f\left(\bx^{-}\right)\right)\right]    \\
&\geq\underset{c^{+}, c^{-} \sim \rho^2}{\mathbb{E}}\underset{\bx\sim\calD_{c^{+}}}{\mathbb{E}}\left[f(x)^T\left(\mu_{c^{+}}-\mu_{c^{-}}\right)\right]\\
&=
(1-\tau) \underset{c^{+}, c^{-} \sim \rho^2}{\mathbb{E}}\left[L_{\text {sup }}^\mu\left(\left\{c^{+}, c^{-}\right\}, f\right) \mid c^{+} \neq c^{-}\right]+\tau\\
&=
(1-\tau) L_{s u p}(f)+\tau
\end{align*}
The bound then follows directly from  Theorem~\ref{th:generalisation bottleneck} and the above results.
\item \textbf{Spectral Contrastive Loss.} We follow the same general idea as in \emph{Case 1}, with changed loss function. Observe that we can bound the spectral by the simple contrastive loss with an additional constant. While this is a very rough bound in this setting we are only interested in bounding the unsupervised by the supervised loss and constants are observed into the big $O$ notation. The bound then follows directly from  Theorem~\ref{th:generalisation bottleneck} and the above results.

\item 
\textbf{Kernel AE.} We can now observe that while the general idea is the same, there is an important difference between the two contrastive approaches above and the kernel AE approach. While in Case 1 and 2 $f(\bx)$ directly gives the embedding function and also a difference between mappings to different classes using positive and negative samples. In the AE case the loss is computed on the reconstruction and not directly on the embedding. In the following we therefore consider $f(\bx)$ as the embedding function. And let us consider the embedding of positive and negative samples as stand ins for the classes.
\begin{align*}
\left|\calL_{u n}(f(\bx)) - \calL_{u n}(f(\bx^-))\right|&\geq\underset{\substack{\bx \sim \calD_{\text {sim }} \\ \bx^{-} \sim \calD_{\text {neg }}}}{\mathbb{E}}\left[f(\bx) - f(\bx_{c^-})\right]    \\
&=\underset{c, c^- \sim \rho^2}{\mathbb{E}}\underset{\bx\sim\calD_{c}}{\mathbb{E}}\left[f(\bx) - f(\bx_{c^-})\right]  \\
&=
(1-\tau) \underset{c, c^- \sim \rho^2}{\mathbb{E}}\left[L_{\text {sup }}^\mu\left(\left\{c, c^-\right\}, f\right) \mid c \neq c^-\right]+\tau\\
&=
(1-\tau) L_{s u p}(f)+\tau
\end{align*}
Again observe that $f(\bx)$ is now only the embedding function, over the class $\mathcal{F}:=\left\{{\bX} \mapsto \bW^T\phi\left({\bX}\right)
:\|\bW^T\|_{\calH} \leq \omega\right\}$. Similarly to the proof of Theorem~\ref{th:generalisation bottleneck} we directly get the complexity term as:
\begin{align*}
    \mathfrak{R}_c(\mathcal{F}, \mathbf{x})\leq \omega\sqrt{h\Tr\left[\bK_{\bX}\right]}.
\end{align*}
Considering it for both positive and negative samples and combined with the above results we obtain the bound for the Kernel AE case. 
\end{enumerate}

This concludes the proof.
\end{proof}

\section{Further Experiments}

In this final section we provide the experimental details for the comparison to neural network methods, referenced in the final section of the main paper as well as the experiments with SVM in addition to $k$-nn as a downstream task. In addition we provide additional experiments on further datasets.

\begin{figure}
    \centering
    \includegraphics[width = \textwidth]{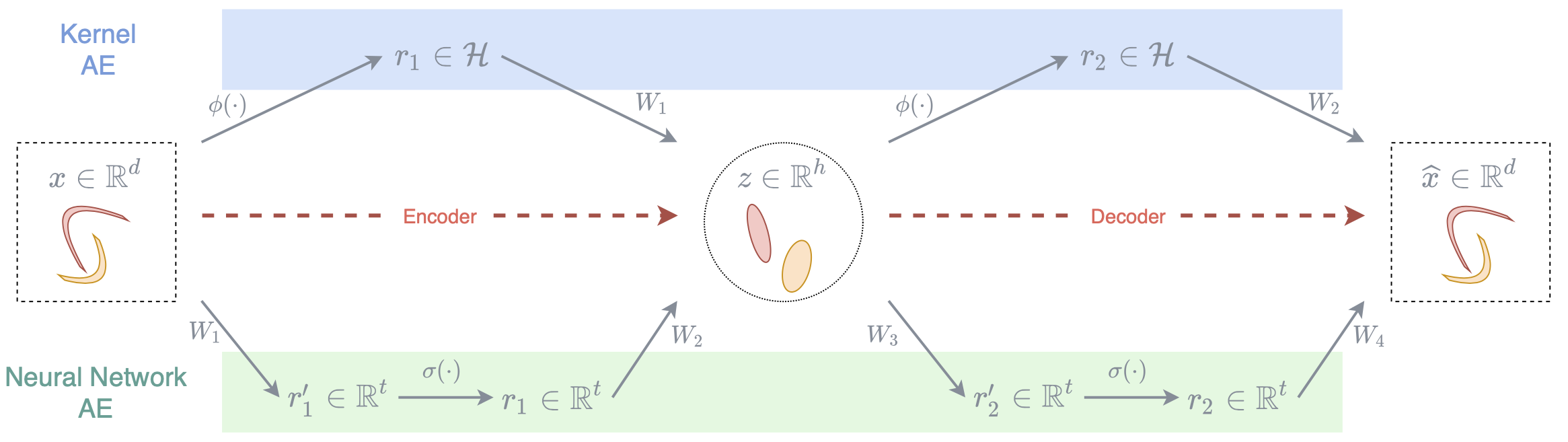}
    \caption{
    Data $\bx$ is mapped to the latent space $\bz$ using the \emph{encoder} and then reconstructed as $\widehat{\bx}$ using the \emph{decoder}. From AE to Kernel AE. 
    \emph{Bottom mapping / green:} Standard fully trained deep leaning approach where hidden layers are mapped into a high dimensional euclidean space.
    \emph{Top mapping / blue:} proposed kernel version where hidden layers are mapped into a Hilbert space.
   }
    \label{fig:illustration bottleneck AE}
\end{figure}

\subsection{Further discussion and experiments comparing neural networks and kernel approaches}\label{sec: relation to deep learning}

As discussed in the introduction, representation learning has become established mainly in the contest of deep learning models. In this paper, we decouple the representation learning paradigm from the widely used deep learning models. While the paper focuses on the specific examples of kernel autoencoders  and kernel contrastive learning, our constructions follow a general principle: instead of considering a (one-hidden layer) neural network $\bW_2\sigma(\bW_1\bx)$  we consider a linear functional in the reproducing kernel Hilbert space $\bW^\top \phi(\bx)$, but still minimise a similar loss functions (reconstruction error in AEs or contrastive losses). We further illustrate this on the example of the Kernel AE.

\textbf{Comparison of deep learning AE and Kernel AE.}
Consider the \emph{kernel AE} illustrated in Figure~\ref{fig:illustration bottleneck AE}. A \emph{deep learning} model is shown in the bottom. The encoder maps the input $\bx\in\bbR^{d}$ to a hidden layer $\br_1 = \sigma(\bW_1\bx) \in \bbR^t$ via a linear map $\bW_1$, and a non-linear activation $\sigma(\cdot)$ with $t\gg d$ and then to a latent representation $\bz = \bW_2\br_1 \in \bbR^k$ with typically $h< d$. Similarly, the decoder maps the representation $\bz$ to the output $\widehat{\bx} = \bW_4\br_2\in \bbR^d$ via the hidden layer $\br_2 = \sigma(\bW_3\bz) \in \bbR^t$. The weights $\bW_1,\ldots,\bW_4$ are learned through a regularized loss minimisation given training samples $\bx_1,\ldots,\bx_n$.
A \emph{non-parametric (kernel) variant} of the AE is obtained by replacing the encoder/decoder with implicit maps $\phi_1: \bbR^d \to \mathcal{H}$, $\phi_2: \bbR^h \to \mathcal{H}$, where $\mathcal{H}$ is the RKHS associated some positive definite kernel $k$.
In the main part we show that, for any new point $\bx^*$, the reconstructed point $\widehat{\bx}^*$ can be expressed only in terms of the kernel evaluation $k(\bx,\bx')$, computed between $\bx^*,\bx_1,\ldots,\bx_n$, \textbf{without explicit knowledge of $\mathcal{H}$, $\phi_1(\cdot)$ or $\phi_2(\cdot)$}.
\begin{figure}
    \centering
    \includegraphics[width = \textwidth]{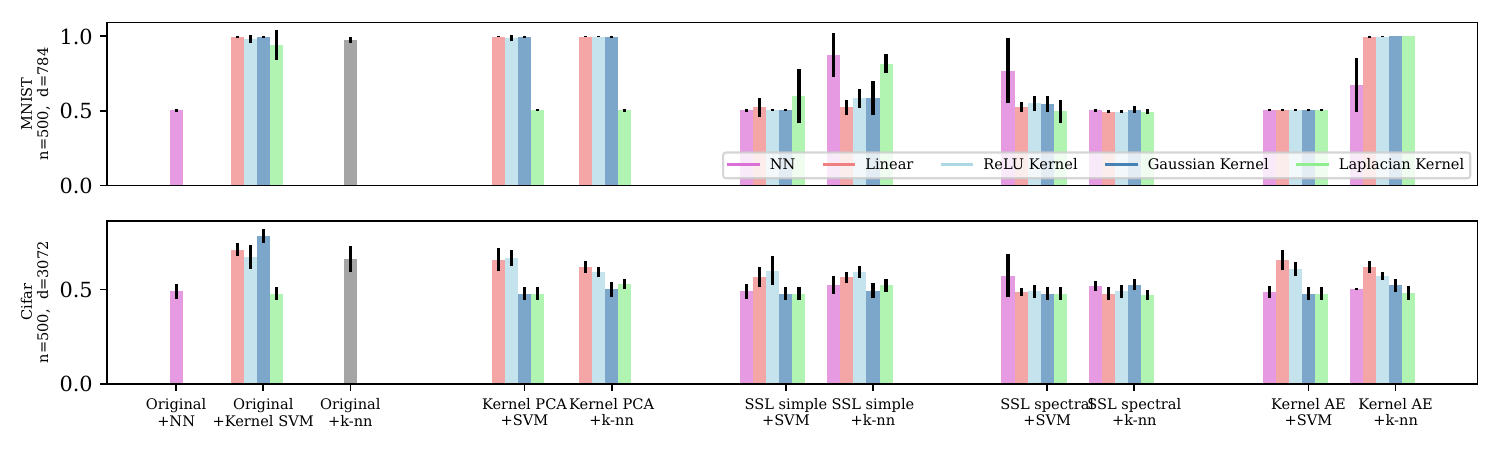}
    \caption{Comparison of kernel methods and neural network models. Plotted is the accuracy for Kernel methods as defined in Section~\ref{sec: experiments} with the addition of a supervised Neural network on the labelled data as well as the neural network models corresponding to the contrastive and AE kernel methods.}
    \label{fig:compare NN app}
\end{figure}

\textbf{Experimental comparison.} Similar to the above comparison we can also define deep learning models analogues to the contrastive SSL models . The general implementation is done in Python with the implementation in \emph{PyTorch} \citep{Pytorch} for fully and optimisation of trained models. The presented setup in Figure~\ref{fig:compare NN app} is the same as in the main paper with some additional experiments. We firstly extend the analysis by considering SVM as a downstream task as well. We can first note that learning a neural network on the small set of labelled data (most left bar) fails, most likely due to the fact that due to the high complexity of the model overfits the training data. 
Overall we observe that under SVM the general comparison between neural network and kernel methods are aligned with the one under $k$-nn.

We will conclude this section with some additional remarks on the comparison and connection between Kernel approaches and neural network methods.

\textbf{Do we need deep kernel representation learning models?} 
Although our construction considers only one a linear functional $\bW^\top \phi(\bx)$, the feature map $\phi(\bx)$ can also capture deeper networks. For instance, one may use $L$-layer ReLU NTK \citep{bietti2021ICLR} in kernel SSL to model the behaviour ``deep'' SSL. 

\textbf{Comparison to generalisation error bounds for deep learning models.}
A common problem in the analysis of modern machine and deep learning methods is that thorough statistical approaches such as VC-dimension \citep{Vapnik_1982_Springer,Vapnik_1998_Wiley} or Rademacher complexity \citep{Tolstikhin_2016_Arxive} do not hold in the overparmeterized learning regime. On the other hand in the context of kernel machines those approaches are well developed \citep{Wahba1990,ScholkopfMIT,BartlettM02}. The proposed extensions allow for a more thorough  theoretical analysis of representation learning as well as kernel variants of unsupervised deep learning methods.

\subsection{Further experiments}

In this section we extend the experiments presented in the main section. We additionally provided the performance of linear SVM as a donwstream classifier in addition to the earlier considered $k$-nn classifier. Furthermore we extend the analysis by considering the following datasets.

We denote the split as $split = unlabelled\%/labelled\%/test\%$.
We show the results for the following three dadatsets:
\emph{concentric circles, factor 0.6} $(n=200,d=2,\#classes=2)$ \citep{scikitlearn},
\emph{cubes} $(n=200,d=13,\#classes=4,split=50/10/40)$ \citep{scikitlearn},
\emph{Iris} $(n=150,d=4,\#classes=3,split=50/5/45)$ \citep{fisher1936use},
\emph{Ionosphere} $(n=351,d=34,\#classes=2,split=50/5/45)$ \citep{misc_ionosphere_52},
\emph{blobs}  $(n=200,d=2,\#classes=3,split=50/5/45)$ \citep{scikitlearn},
\emph{Breast cancer}  $(n=569,d=30,\#classes=2,split=50/5/45)$ \cite{misc_breast_cancer_wisconsin},
\emph{Heart Failure} $(n=299,d=13,\#classes=2,split=50/5/45)$ \cite{misc_heart_failure_clinical_records_519},
\emph{Mushroom (sub-sample)} $(n=200,d=22,\#classes=2,split=50/5/45)$  \cite{misc_mushroom_73},
\emph{Wine} $(n=178,d=13,\#classes=3,split=50/10/4)$  \cite{misc_wine_109},
\emph{Parkinson's Disease Classification from Speech} $(n=756,d=754,\#classes=2,split=50/5/45)$   \cite{pd_dataset} and 
\emph{Moons} $(n=200,d=2,\#classes=2,split=50/5/45)$ \citep{scikitlearn}.

We can conclude this section by observing that overall the main findings in the main paper stay consistent throughout the analysis of additional datasets and downstream tasks. 

\begin{figure}[t]
    \centering
    \includegraphics[width = \textwidth]{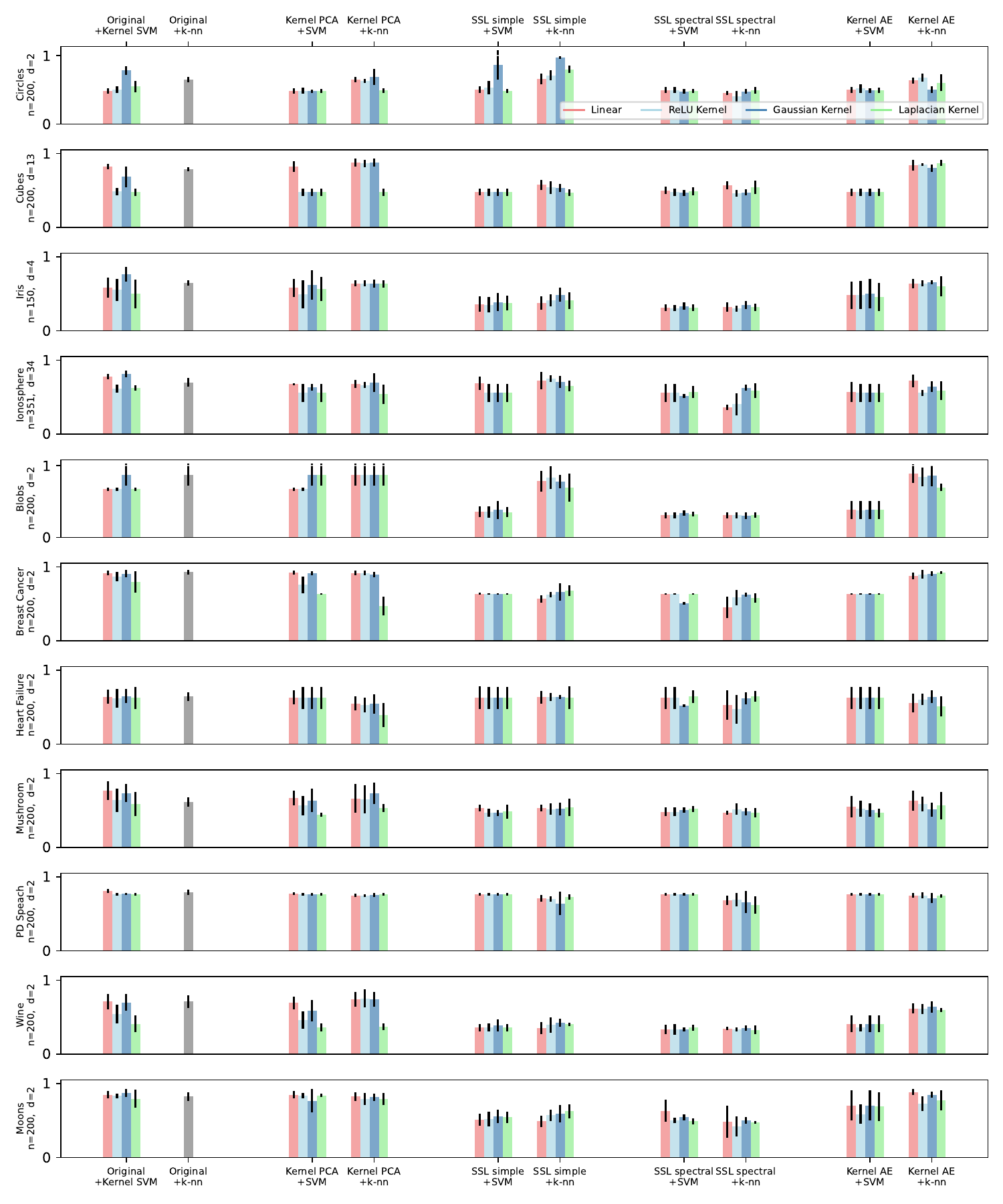}
    \caption{Accuracy for different methods. From left to right: we first consider $k$-nn and Kernel SVM on the original features followed by SVM and $k$-nn on embeddings obtained by  Kernel SVM, simple contrastive kernel method, spectral contrastive kernel method and kernel AE.
    }
    \label{fig: main experiments app}
\end{figure}